\documentclass[11pt]{article}
\usepackage[margin=1.1in]{geometry}

\usepackage{amsmath}       
\usepackage{amsfonts}       
\usepackage{amsthm}
\usepackage{mathtools}
\usepackage{shortcuts}

\usepackage{cases}
\usepackage{float}
\usepackage[utf8]{inputenc} 
\usepackage[T1]{fontenc}    
\usepackage{hyperref}
\usepackage{url}            
\usepackage{booktabs}       
\usepackage{nicefrac}       
\usepackage{microtype}      
\usepackage[toc,page]{appendix}

\usepackage{graphicx}
\graphicspath{{figs/}}
\usepackage{svg}
\usepackage{caption}
\usepackage{subfig}
\usepackage[export]{adjustbox}
\usepackage{wrapfig}
\usepackage{xcolor}
\usepackage{stmaryrd}
\usepackage[percent]{overpic}
\usepackage{authblk}
\usepackage{tabularx}

\begin{document}

\title{Uncovering the structure of clinical EEG signals with self-supervised learning}
\date{\vspace{-5ex}}

\author[1,2]{Hubert Banville\thanks{hubert.jacob-banville@inria.fr}}
\author[1]{Omar Chehab}
\author[1,3]{Aapo Hyv\"arinen}
\author[1,4]{Denis-Alexander Engemann}
\author[1]{Alexandre Gramfort}
\affil[1]{Université Paris-Saclay, Inria, CEA, Palaiseau, France}
\affil[2]{InteraXon Inc., Toronto, Canada}
\affil[3]{Dept. of CS and HIIT, University of Helsinki, Finland}
\affil[4]{Max Planck Institute for Human Cognitive and Brain Sciences, Department of Neurology, Leipzig, Germany}


\maketitle
\begin{abstract}
\textit{Objective.}
Supervised learning paradigms are often limited by the amount of labeled data that is available.
This phenomenon is particularly problematic in clinically-relevant data, such as electroencephalography (EEG), where labeling can be costly in terms of specialized expertise and human processing time.
Consequently, deep learning architectures designed to learn on EEG data have yielded relatively shallow models and performances at best similar to those of traditional feature-based approaches. 
However, in most situations, unlabeled data is available in abundance.
By extracting information from this unlabeled data, it might be possible to reach competitive performance with deep neural networks despite limited access to labels.
\textit{Approach.}
We investigated self-supervised learning (SSL), a promising technique for discovering structure in unlabeled data, to learn representations of EEG signals.
Specifically, we explored two tasks based on temporal context prediction as well as contrastive predictive coding on two clinically-relevant problems: EEG-based sleep staging and pathology detection.
We conducted experiments on two large public datasets with thousands of recordings and performed baseline comparisons with purely supervised and hand-engineered approaches.
\textit{Main results.}
Linear classifiers trained on SSL-learned features consistently outperformed purely supervised deep neural networks in low-labeled data regimes while reaching competitive performance when all labels were available.
Additionally, the embeddings learned with each method revealed clear latent structures related to physiological and clinical phenomena, such as age effects. 
\textit{Significance.}
We demonstrate the benefit of self-supervised learning approaches on EEG data.
Our results suggest that SSL may pave the way to a wider use of deep learning models on EEG data.
\end{abstract}

\noindent{\it Keywords\/}
Self-supervised learning, representation learning, machine learning, electroencephalography, sleep staging, pathology detection, clinical neuroscience

\section{Introduction}

Electroencephalography (EEG) and other biosignal modalities have enabled numerous applications inside and outside of the clinical domain, e.g., studying sleep patterns and their disruption \cite{ghassemi2018you}, monitoring seizures \cite{acharya2013automated} and brain-computer interfacing \cite{lotte2018review}.
In the last few years, the availability and portability of these devices has increased dramatically, effectively democratizing their use and unlocking the potential for positive impact on people's lives \cite{mihajlovic2015wearable,casson2019wearable}.
For instance, applications such as at-home sleep staging and apnea detection, pathological EEG detection, mental workload monitoring, etc. are now entirely possible.

In all these scenarios, monitoring modalities generates an ever-increasing amount of data which needs to be interpreted.
Therefore, predictive models that can classify, detect and ultimately ``understand'' physiological data are required.
Traditionally, this type of modelling has mostly relied on supervised approaches, where large datasets of annotated examples are required to train models with high performance.

However, obtaining accurate annotations on physiological data can prove expensive, time consuming or simply impossible.
For example, annotating sleep recordings requires trained technicians to go through hours of data visually and label 30-s windows one-by-one \cite{malhotra2013sleep}.
Clinical recordings such as those used to diagnose epilepsy or brain lesions must be reviewed by neurologists, who might not always be available.
More broadly, noise in the data and the complexity of brain processes of interest can make it difficult to interpret and annotate EEG signals, which can lead to high inter-rater variability, i.e., label noise \cite{younes2016staging,engemann018robust}.
Furthermore, in some cases, knowing exactly what the participants were thinking or doing in cognitive neuroscience experiments can be challenging, making it hard to obtain accurate labels. 
In imagery tasks, for instance, the subjects might not be following instructions or the process under study might be difficult to quantify objectively (e.g., meditation, emotions).
Therefore, a new paradigm that does not rely primarily on supervised learning is necessary for making use of large unlabeled sets of recordings such as those generated in the scenarios described above.
However, traditional unsupervised learning approaches such as clustering and latent factor models do not offer fully satisfying answers as their performance is not as straightforward to quantify and interpret as supervised ones.

``Self-supervised learning'' (SSL) is an unsupervised learning approach that learns representations from unlabeled data, exploiting the structure of the data to provide supervision \cite{jing2019self}.
By reframing an unsupervised learning problem as a supervised one, SSL allows the use of standard, better understood optimization procedures.
SSL comprises a ``pretext'' and a ``downstream'' task.
The downstream task is the task one is actually interested in but for which there are limited or no annotations.
The pretext task, on the other hand, must be sufficiently related to the downstream task such that similar representations should be employed to carry it out; importantly, it must be possible to generate the annotations for this pretext task using the unlabeled data alone.
For example, in a computer vision scenario, one could use a jigsaw puzzle task where patches are extracted from an image, scrambled randomly and then fed to a neural network that is trained to recover the original spatial ordering of the patches \cite{noroozi2016unsupervised}.
If the network performs this task reasonably well, then it is conceivable that it has learned some of the structure of natural images, and that the trained network could be reused as a feature extractor or weight initialization on a smaller-scale supervised learning problem such as object recognition.
Apart from facilitating the downstream task and/or reducing the number of necessary annotated examples, self-supervision can also uncover more general and robust features than those learned in a specialized supervised task \cite{oord2018representation}.
Therefore, given the potential benefits of SSL, can it be used to enhance the analysis of EEG?

To date, most applications of SSL have focused on domains where plentiful annotated data is already available, such as computer vision \cite{jing2019self} and natural language processing \cite{mikolov2013efficient,devlin2018bert}.
Particularly in computer vision, deep networks are often trained with fully supervised tasks (e.g., ImageNet pretraining). In this case, enough labeled data is available such that direct supervised learning on the downstream task is already in itself competitive \cite{he2019rethinking}.
SSL has an even greater potential in domains where low-labeled data regimes are common and supervised learning's effectiveness is limited, e.g. biosignal and EEG processing.
Despite this, few studies on SSL and biosignals have been published.
These studies either focus on limited downstream tasks and datasets \cite{yuan2017wave}, or test their approach on signals other than EEG \cite{sarkar2020self}.

Therefore, it still remains to be shown whether self-supervision can truly bring improvements over standard supervised approaches on EEG, and if this is the case, what the best ways of applying it are.
Specifically, can we learn generic representations of EEG with self-supervision and, in doing so, reduce the need for costly EEG annotations?
Given the growing popularity of deep learning as a processing tool for EEG \cite{roy2019deep}, the answer could have a significant impact on current practices in the field of EEG processing.
Indeed, while deep learning is notoriously data-hungry, an overwhelmingly large part of all neuroscience research happens in the low-labeled data regime, including EEG research: clinical studies with a few hundred subjects are often considered to be big data, while large-scale studies are much rarer and usually originate from research consortia \cite{engemann2020combining,obeid2016temple,bycroft2018uk,shafto2014cambridge}.
Therefore, it is to be expected that the performance reported by most deep learning-EEG studies - often in low-labeled data regimes - has so far remained limited and does not clearly outperform those of conventional approaches \cite{roy2019deep}.
By leveraging unlabeled data, SSL can effectively create a lot more examples, which could enable more successful applications of deep learning to EEG.

In this paper, we investigate the use of self-supervision as a general approach to learning representations from EEG data.
To the best of our knowledge, we present the first detailed analysis of SSL tasks on multiple types of EEG recordings.
We aim to answer the following questions:
\begin{enumerate}
    \item What are good SSL tasks that capture relevant structure in EEG data?
    \item How do SSL features compare to other unsupervised and supervised approaches in terms of downstream classification performance? 
    \item What are the characteristics of the features learned by SSL? Specifically, can SSL capture physiologically- and clinically-relevant structure from unlabeled EEG?
\end{enumerate}


The rest of the paper is structured as follows.
Section~\ref{sec:methods} presents an overview of the SSL literature, then describes the different SSL tasks and learning problems considered in our study. 
We also introduce the neural architectures, baseline methods and data used in our experiments.
Next, Section~\ref{sec:experiments} reports the results of our experiments on EEG.
Lastly, we discuss the results in Section~\ref{sec:discussion}.

\section{Methods}
\label{sec:methods}

\subsection{State-of-the-art self-supervised learning approaches}

Although it has not always been known as such, SSL has already been used in many other fields.
In computer vision, multiple approaches have been proposed that rely on the spatial structure of images and the temporal structure of videos.
For example, a context prediction task was used to train feature extractors on unlabeled images in \cite{doersch2015unsupervised} by predicting the position of a randomly sampled image patch relative to a second patch.
Using this approach to pretrain a neural network, the authors reported improved performance as compared to a purely supervised model on the Pascal VOC object detection challenge.
These results were among the first showing that self-supervised pretraining could help improve performance when limited annotated data is available.
Similarly, the jigsaw puzzle task mentioned above \cite{noroozi2016unsupervised} led to improved downstream performance on the same dataset.
In the realm of video processing, approaches based on temporal structure have also been proposed: for instance, in \cite{misra2016shuffle}, predicting whether a sequence of video frames were ordered or shuffled was used as a pretext task and tested on a human activity recognition downstream task.
The interested reader can find other applications of SSL to images in \cite{jing2019self}.

Similarly, modern natural language processing (NLP) tasks often rely on self-supervision to learn word embeddings, which are at the core of many applications \cite{turian2010word}.
For instance, the original \textit{word2vec} model was trained to predict the words around a center word or a center word based on the words around it \cite{mikolov2013efficient}, and then reused on a variety of downstream tasks \cite{nayak2016evaluating}.
More recently, a dual-task self-supervised approach, \textit{BERT}, led to state-of-the-art performance on 11 NLP tasks such as question answering and named entity recognition \cite{devlin2018bert}.
The high performance achieved by this approach showcases the potential of SSL for learning general-purpose representations.

Lately, more general pretext tasks as well as improved methodology have led to strong results that have begun to rival purely supervised approaches.
For instance, contrastive predictive coding (CPC), an autoregressive prediction task in latent space, was successfully used for images, text and speech~\cite{oord2018representation}.
Given an encoder and an autoregressive model, the task consists of predicting the output of the encoder for future windows (or image patches or words) given a context of multiple windows.
The authors presented several improved results on various downstream tasks; a follow-up further showed that higher-capacity networks could improve downstream performance even more, especially in low-labeled data regimes \cite{henaff2019data}.
Momentum contrast (MoCo), rather than proposing a new pretext task, is an improvement upon contrastive tasks, i.e., where a classifier must predict which of two or more inputs is the true sample \cite{he2019momentum,chen2020improved}.
By improving the sampling of negative examples in contrastive tasks, MoCo helped boost the efficiency of SSL training as well as the quality of the representations learned.
Similarly, it was found in \cite{chen2020simple} that using the right data augmentation transforms (e.g., random cropping and color distortion on images) and increasing batch size could lead to significant improvements in downstream performance.

The ability of SSL-trained features to demonstrably generalize to downstream tasks justifies a closer look at their statistical structure. 
A general and theoretically grounded approach was recently formalized by Hyv\"arinen \textit{et al.} \cite{hyva17aistats, hyvarinen2019nonlinear} from the perspective of nonlinear independent components analysis.
In this generalized framework, an observation $\mathbf{x}$ is embedded using an invertible neural network, and contrasted against an auxiliary variable $\mathbf{u}$ (e.g., the time index, the index of a segment or the history of the data). 
A discriminator classifies the pair by learning to predict whether $\mathbf{x}$ is paired with its corresponding auxiliary variable $\mathbf{u}$ or a perturbed (random) one $\mathbf{u^*}$.
When the data exhibits certain structure (e.g., autocorrelation, non-stationarity, non-gaussianity), the embedder trained on this contrastive task will perform identifiable nonlinear ICA \cite{hyvarinen2019nonlinear}.
Most of the previously introduced SSL tasks can be viewed through this framework.
Given the widespread use of linear ICA as a preprocessing and feature extraction tool in the EEG community \cite{makeig1997blind,jung1998extended,parra2005recipes,ablin2018faster}, an extension to the nonlinear regime is a natural step forward and could help improve traditional processing pipelines.

Remarkably, very few studies have applied SSL to biosignals despite its potential to leverage large quantities of unlabeled data.
In \cite{yuan2017wave}, a model inspired by \textit{word2vec}, called \textit{wave2vec}, was developed to work with EEG and electrocardiography (ECG) time series.
Representations were learned by predicting the features of neighbouring windows from the concatenation of time-frequency representations of EEG signals and demographic information.
This approach was however only tested on a single EEG dataset and was not benchmarked against fully supervised deep learning approaches or expert feature classification.
SSL has also been applied to ECG as a way to learn features for a downstream emotion recognition task: in \cite{sarkar2020self}, a transformation discrimination pretext task was used in which the model had to predict which transformations had been applied to the raw signal.
While these results show the potential of self-supervised learning for biosignals, a more extensive analysis of SSL targeted at EEG is required to pave the way for practical applications.

\subsection{Self-supervised learning pretext tasks for EEG}
\label{subsec:ssl}

\begin{figure}
    \centering
    \includegraphics[width=\textwidth]{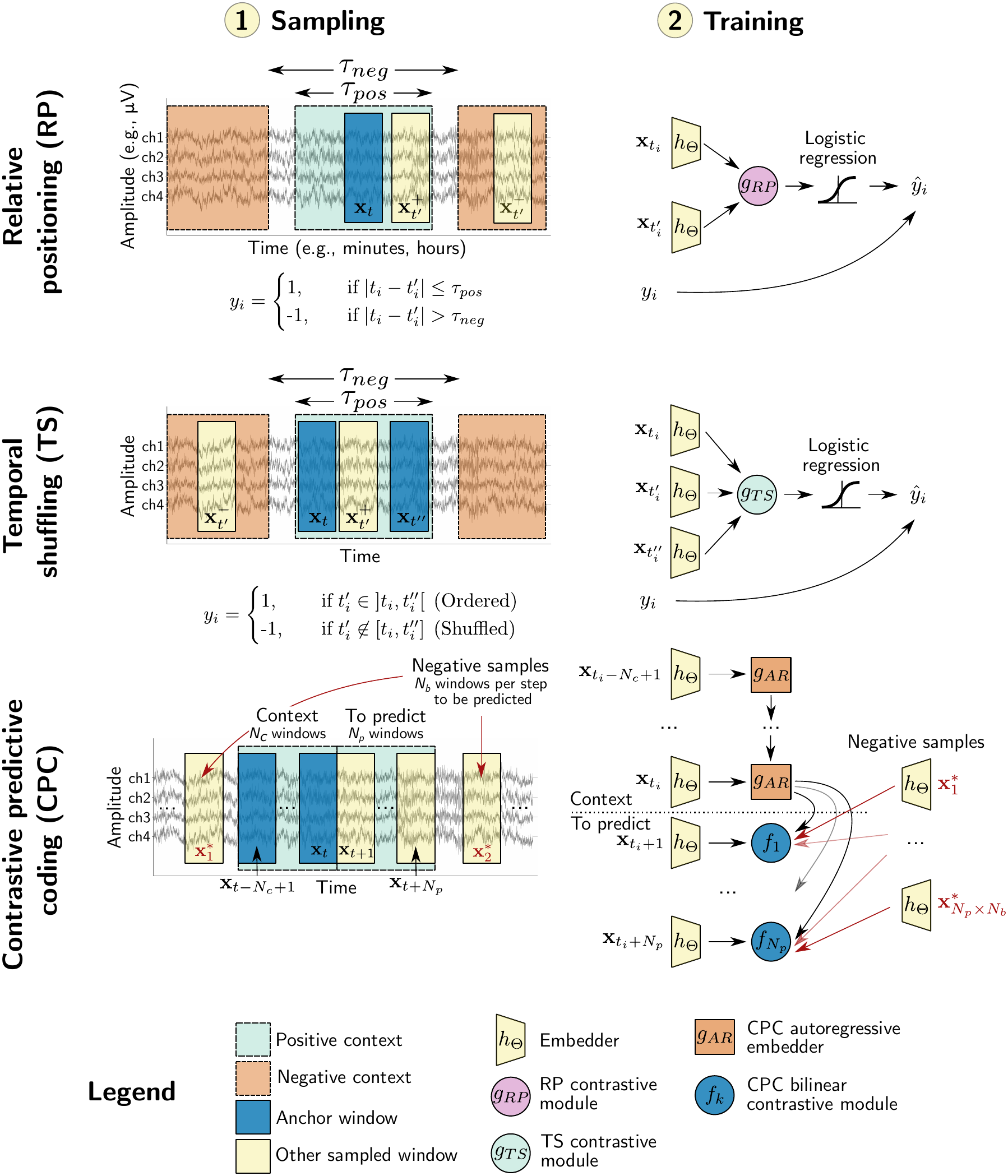}
    \caption{Visual explanation of the three SSL pretext tasks used in this study. The first column illustrates the sampling process by which examples are obtained in each pretext task. The second column describes the training process, where sampled examples are used to train a feature extractor $h_\Theta$ end-to-end.}
    \label{fig:pretext_tasks}
\end{figure}

In this section, we describe the three SSL pretext tasks used in the paper.
A visual explanation of the tasks can be found in Fig.~\ref{fig:pretext_tasks}.

\paragraph*{Notation}
We denote by $\intset{q}$ the set $\{1, \ldots, q\}$ and by $\intset{p,q}$  the set $\{p, \ldots, q\}$ for any integer $p, q \in \bbN$.
The index $t$ refers to time indices in the multivariate time series $S \in \mathbb{R}^{C \times M}$, where $M$ is the number of time samples and $C$ is the dimension of samples (\eg channels).
We assume for simplicity that each $S$ has the same size. We denote by $y \in \{-1, 1\}$ a binary label used in the learning task.

\subsubsection{Relative positioning}

To produce labeled samples from the multivariate time series $S$, we propose to sample pairs of time windows $(x_{t}, x_{t'})$ where each window $x_t$, $x_{t'}$ is in $\bbR^{C \times T}$ and $T$ is the duration of each window, and where the index $t$ indicates the time sample at which the window starts in $S$.
The first window $x_{t}$ is referred to as the ``anchor window''.
Our assumption is that an appropriate representation of the data should evolve slowly over time (akin to the driving hypothesis behind Slow Feature Analysis (SFA) \cite{becker1993learning,wiskott2002slow}) suggesting that time windows close in time should share the same label.
In the context of sleep staging, for instance, sleep stages usually last between 1 to 40 minutes \cite{altevogt2006sleep}; therefore, nearby windows likely come from the same sleep stage, whereas faraway windows likely come from different sleep stages.
Given $\tau_{pos} \in \bbN$, which controls the duration of the positive context, and $\tau_{neg} \in \bbN$, which corresponds to the negative context around each window $x_i$, we sample $N$ labeled pairs:
\begin{equation*}
\mathcal{Z}_N = \{ ((x_{t_i}, x_{t_i'}), y_i)~|~i \in \intset{N}, (t_i, t'_i) \in \cT, y_i \in \cY\},
\end{equation*}
where $\cY = \{-1, 1\}$ and $\cT = \{(t,t') \in \intset{M - T + 1}^2 ~|~ \left|t - t'\right| \leq \tau_{pos} \textrm{\ or\ } \left|t - t'\right| > \tau_{neg}\}$.
Intuitively, $\cT$ is the set of all pairs of time indices $(t,t')$ which can be constructed from windows of size $T$ in a time series of size $M$, given the duration constraints imposed by the particular choices of $\tau_{pos}$ and $\tau_{neg}$\footnote{The values of $\tau_{pos}$ and $\tau_{neg}$ can be selected based on prior knowledge of the signals and/or with a hyperparameter search.}.
Here $y_i \in \cY$ is specified by the positive or negative contexts parameters:
\begin{numcases}{y_{i} =}
    1, & if $|t_i-t_i'|\leq\tau_{pos}$ \nonumber \\
    -1, & if $|t_i-t_i'|>\tau_{neg}$
\end{numcases}.

We ignore window pairs where $x_{t'}$ falls outside of the positive and negative contexts of the anchor window $x_t$.
In other words, the label indicates whether two time windows are closer together than $\tau_{pos}$ or farther apart than $\tau_{neg}$ in time.
Noting the connection with the task in \cite{doersch2015unsupervised}, we call this pretext task ``relative positioning'' (RP).

In order to learn end-to-end how to discriminate pairs of time windows based on their relative position, we introduce two functions $h_\Theta$ and $g_{RP}$. 
$h_\Theta: \bbR^{C\times T} \rightarrow \bbR^D$ is a feature extractor with parameters $\Theta$ which maps a window $x$ to its representation in the feature space.
Ultimately, we expect $h_\Theta$ to learn an informative representation of the raw EEG input which can be reused in different downstream tasks.
A contrastive module $g_{RP}$ is then used to aggregate the feature representations of each window.
For the RP task, $g_{RP}: \bbR^D \times \bbR^D \rightarrow \bbR^D$ combines representations from pairs of windows by computing an elementwise absolute difference, denoted by the $|\cdot|$  operator: $g_{RP}(h_\Theta(x), h_\Theta(x')) = |h_\Theta(x) - h_\Theta(x')| \in \bbR^D$.
The role of $g_{RP}$ is to aggregate the feature vectors extracted by $h_\Theta$ on the two input windows and highlight their differences to simplify the contrastive task.
Finally, a linear context discriminative model with coefficients $w \in \bbR^D$ and bias term $w_0 \in \bbR$ is responsible for predicting the associated target $y$.
Using the binary logistic loss on the predictions of $g_{RP}$ we can write a joint loss function $\mathcal{L}(\Theta,w,w_0)$ as
\begin{equation} \label{eq:loss_rp}
\mathcal{L}(\Theta,w,w_0) = \\
  \smashoperator{\sum_{(x_t, x_{t'}, y) \in \mathcal{Z}_N}}
    \log(1 + \exp(-y [w^\top g_{RP}(h_\Theta(x_t), h_\Theta(x_{t'})) + w_0])) 
\end{equation}
which we assume to be fully differentiable with respect to the parameters $(\Theta,w,w_0)$. Given
the convention used for $y$, the predicted target is the sign of $w^\top g(h_\Theta(x_t), h_\Theta(x_{t'})) + w_0$.

\subsubsection{Temporal shuffling}

We also introduce a variation of the RP task that we call ``temporal shuffling'' (TS), in which we instead sample two anchor windows $x_{t}$ and $x_{t''}$ from the positive context, and a third window $x_{t'}$ that is either between the first two windows or in the negative context.
We then construct window triplets that are either temporally ordered ($t < t' < t''$) or shuffled ($t < t'' < t'$ or $t' < t < t''$).
We augment the number of possible triplets by also considering the mirror image of the previous triplets, e.g., $(x_t, x_{t'}, x_{t''})$ becomes $(x_{t''}, x_{t'}, x_{t})$.
The label $y_i$ then indicates whether the three windows are ordered or have been shuffled, similar to \cite{misra2016shuffle}.

The contrastive module for TS is defined as $g_{TS}: \bbR^D \times \bbR^D \times \bbR^D \rightarrow \bbR^{2D}$ and is implemented by concatenating the absolute differences:
$$
g_{TS}(h_\Theta(x), h_\Theta(x'), h_\Theta(x'')) = (\abs{h_\Theta(x) -h_\Theta(x')}, \abs{h_\Theta(x')-h_\Theta(x'')}) \in \bbR^{2D} \enspace .
$$
Moreover, Eq.~\eqref{eq:loss_rp} is extended to TS by replacing $g_{RP}$ by $g_{TS}$ and introducing $x_{t''}$ to obtain:
\begin{equation} \label{eq:loss_ts}
\mathcal{L}(\Theta,w,w_0) = \\
  \smashoperator{\sum_{(x_t, x_{t'}, x_{t''}, y) \in \mathcal{Z}_N}}
    \log(1 + \exp(-y [w^\top g_{TS}(h_\Theta(x_t), h_\Theta(x_{t'}), h_\Theta(x_{t''})) + w_0])) \enspace .
\end{equation}

\subsubsection{Contrastive predictive coding}
\label{subsec:cpc}

The contrastive predictive coding (CPC) pretext task, introduced by Oord et al. \cite{oord2018representation}, is defined here in comparison to RP and TS, as all three tasks share key similarities.
Indeed, CPC can be seen as an extension of RP, where the single anchor window $x_t$ is replaced by a sequence of $N_c$ non-overlapping windows that are summarized by an autoregressive encoder $g_{AR}: \bbR^{D \times  N_c} \rightarrow \bbR^{D_{AR}}$ with parameters $\Theta_{AR}$\footnote{
CPC's encoder $g_{AR}$ has parameters $\Theta_{AR}$, however we omit them from the notation for brevity.}.
This way, the information in the context can be represented by a single vector $c_t \in \bbR^{D_{AR}}$.
$g_{AR}$ can be implemented for example as a recurrent neural network with gated-recurrent units (GRU).

The context vector $c_t$ is paired with not one, but $N_p$ future windows (or ``steps'') which immediately follow the context.
Negative windows are then sampled in a similar way as with RP and TS when $\tau_{neg}=0$, i.e., the negative context is relaxed to include the entire time series.
For each future window, $N_b$ negative windows $x^*$ are sampled inside each multivariate time series $S$ (``same-recording negative sampling'') or across all available $S$ (``across-recording negative sampling'').
For the sake of simplicity and to follow the notation of the original CPC article, we modify our notation slightly: we now denote a time window by $x_t$ where $t$ is the index of the window in the list of all non-overlapping windows of size $T$ that can be extracted from a time series $S$.
Therefore, the procedure for building a dataset with $N$ examples boils down to sampling sequences $X^c$, $X^p$ and $X^n$ in the following manner:
%
%
\begin{align*}
X^{c}_{i} &= (x_{t_i-N_c+1}, \ldots, x_{t_i}) \quad&\text{($N_c$ context windows)} \\
X^{p}_{i} &= (x_{t_i + 1}, \ldots, x_{t_i + N_p}) \quad&\text{($N_p$ future windows)}\\
X^{n}_{i} &= (x_{t^*_{i_{1, 1}}}, \ldots, x_{t^*_{i_{1, N_b}}}, \ldots, x_{t^*_{i_{N_p, 1}}}, \ldots, x_{t^*_{i_{N_p, N_b}}}) \quad&\text{($N_p N_b$ random negative windows)}
\end{align*}
where $t_i \in \intset{N_c, M-N_p}$. We denote with $t^*$ time indices of windows sampled uniformly at random. The dataset then reads:
\begin{equation}
\mathcal{Z}_N = \{ (X^{c}_{i}, X^{p}_{i}, X^{n}_{i}) ~|~i \in \intset{N}\} \enspace .
\end{equation}
%
%





As with RP and TS, the feature extractor $h_\Theta$ is used to extract a representation of size $D$ from a window $x_t$.
Finally, whereas the contrastive modules $g_{RP}$ and $g_{TS}$ explicitly relied on the absolute value of the difference between embeddings $h$, here for each future window $x_{t+k}$ where ${k\in\intset{N_p}}$ a bilinear model $f_k$ parametrized by $W_k \in \bbR^{D \times D_{AR}}$ is used to predict whether the window chronologically follows the context $c_t$ or not:
\begin{equation}
f_k(c_t, h_\Theta\textbf{}(x_{t+k})) = h_\Theta(x_{t+k})^\top W_k c_t
\end{equation}

The whole CPC model is trained end-to-end using the InfoNCE loss \cite{oord2018representation} (a categorical cross-entropy loss) defined as
%
%
\begin{align} \label{eq:loss_cpc}
\begin{split}
\mathcal{L}(\Theta,&\Theta_{AR},W_k,\ldots,W_{k+N_p-1}) = \\
    &-\sum_{\substack{(X_i^c, X_i^p, X_i^n) \in \mathcal{Z}_N \\ c_{t_i} = g_{AR}(X_i^c)}} \sum_{k=1}^{N_p} \log \left[ \frac{\exp (f_k(c_{t_i},h_\Theta(x_{t_i+k})))}{\exp(f_k(c_{t_i},h_\Theta(x_{t_i+k}))) + \smashoperator{\sum_{j \in \intset{N_b}}}{\exp(f_k(c_{t_i},h_\Theta(x_{t^*_{i_{k, j}}})))}}\right]
\end{split}
\end{align}

While in RP and TS the model must predict \textit{whether} a pair is positive or negative, in CPC the model must pick \textit{which} of $N_b + 1$ windows actually follows the context.
In practice, we sample batches of $N_b+1$ sequences and for each sequence use the $N_b$ other sequences in the batch to supply negative examples.

\subsection{Downstream tasks}

We performed empirical benchmarks of EEG-based SSL on two clinical problems that are representative of the current challenges in machine learning-based analysis of EEG: sleep monitoring and pathology screening.
These two clinical problems commonly give rise to classification tasks, albeit with different numbers of classes and distinct data-generating mechanisms: sleep monitoring is concerned with biological events (event level) while pathology screening is concerned with single patients as compared to the population (subject level).
These two clinical problems have generated considerable attention in the research community, which has led to the curation of large public databases. 
To enable fair comparisons with supervised approaches, we benchmarked SSL on the Physionet Challenge 2018 \cite{ghassemi2018you, goldberger2000physiobank} and the TUH Abnormal EEG \cite{lopez2017automated} datasets.

First, we considered sleep staging, which is a critical component of a typical sleep monitoring assessment and is key to diagnosing and studying sleep disorders such as apnea and narcolepsy \cite{bathgate2019diagnostic}. 
Sleep staging has been extensively studied in the machine (and deep) learning literature \cite{chambon2018deep,motamedi2014signal,roy2019deep} (approximately 10\% of reviewed papers in \cite{roy2019deep}), though not through the lens of SSL.
Achieving fully automated sleep staging could have a substantial impact on clinical practice as (1) agreement between human raters is often limited \cite{younes2016staging} and (2) the annotation process is time-consuming and still largely manual \cite{malhotra2013sleep}.
Sleep staging typically gives rise to a 5-class classification problem where the possible predictions are W (wake), N1, N2, N3 (different levels of sleep) and R (rapid eye movement periods).
Here, the task consists of predicting the sleep stages that correspond to 30-s windows of EEG.

Second, we applied SSL to pathology detection: EEG is routinely used in a clinical context to screen individuals for neurological conditions such as epilepsy and dementia \cite{smith2005eeg,micanovic2014diagnostic}.
However, successful pathology detection requires highly specialized medical expertise and its quality depends on the expert's training and experience. 
Automated pathology detection could, therefore, have a major impact on clinical practice by facilitating neurological screening. 
This gives rise to classification tasks at the subject level where the challenge is to infer the patient's diagnosis or health status from the EEG recording. 
In the TUH dataset, medical specialists have labeled recordings as either pathological or non-pathological, giving rise to a binary classification problem.
Importantly, these two labels reflect highly heterogeneous situations: a pathological recording could reflect anomalies due to various medical conditions, suggesting a rather complex data-generating mechanism.
Again, various supervised approaches, some of them leveraging deep architectures, have addressed this task in the literature \cite{lopez2015automated,tibor2017deep,gemein2020machine}, although none has relied on self-supervision.

These two tasks are further described in Section~\ref{subsec:data} when discussing the data used in our experiments.

\subsection{Deep learning architectures} 

\begin{figure}
    \centering
    \includegraphics[width=\textwidth]{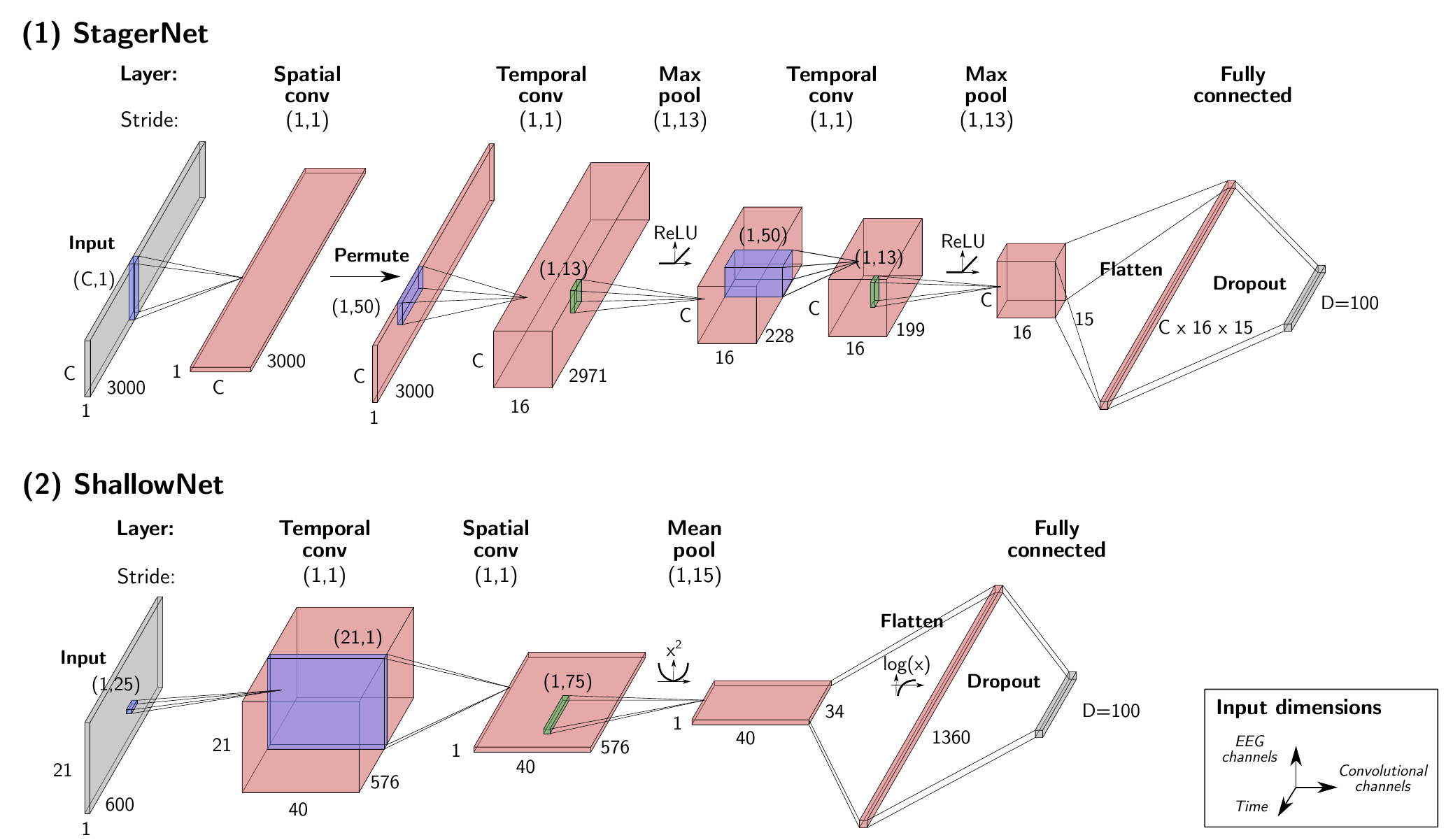}
    \caption{Neural network architectures used as embedder $h_\Theta$ for (1) sleep EEG and (2) pathology detection experiments.}
    \label{fig:architectures}
\end{figure}

We used two different deep learning architectures as embedders $h_\Theta$ in our experiments (see Fig.~\ref{fig:architectures} for a detailed description).
Both architectures were convolutional neural networks composed of spatial and temporal convolution layers, which respectively learned to perform the spatial and temporal filtering operations typical of EEG processing pipelines.

The first one, which we call StagerNet, was adapted from previous work on sleep staging where it was shown to perform well for window-wise classification of sleep stages \cite{chambon2018deep}.
StagerNet is a 3-layer convolutional neural network optimized to process windows of 30\,s of multichannel EEG.
As opposed to the original architecture, (1) we used twice as many convolutional channels (16 instead of 8), (2) we added batch normalization after both temporal convolution layers\footnote{As described in \cite{oord2018representation,he2019momentum}, batch normalization can harm the network's ability to learn on the CPC pretext task. However, we did not see this effect on our models (likely because their capacity is relatively small) and alternatives such as no normalization or layer normalization \cite{ba2016layer} performed unfavorably. Therefore, we also used batch normalization in CPC experiments.} (3) we did not pad temporal convolutions and (4) we changed the dimensionality of the output layer to $D=100$ instead of the number of classes (see Fig.~\ref{fig:architectures}-1).
This yielded a total of 62,307 trainable parameters.

The second embedder architecture, ShallowNet, was directly taken from previous literature on the TUH Abnormal dataset \cite{tibor2017deep, gemein2020machine}.
Originally designed to be a parametrized version of the filter bank common spatial patterns (FBCSP) processing pipeline common in brain-computer interfacing, ShallowNet has a single (split) convolutional layer followed by a squaring non-linearity, average pooling, a logarithm non-linearity, and a linear output layer.
Batch normalization was used after the temporal convolution layer.
Despite its simplicity, this architecture was shown in \cite{gemein2020machine} to perform almost as well as the best model on the task of pathology detection on the TUH Abnormal dataset.
We therefore used it as is, except for the dimensionality of the output layer which we also changed to $D=100$ (See Fig.~\ref{fig:architectures}-2).
This yielded a total of 170,860 trainable parameters.

We used a GRU with a hidden layer of size $D_{AR}=100$ for the CPC task's $g_{AR}$, for experiments on both datasets.

The Adam optimizer \cite{kingma2014adam} with $\beta_1=0.9$ and $\beta_2=0.999$ and learning rate $5\times10^{-4}$ was used.
The batch size for all deep models was set to 256, except for CPC where it was set to 32.
Training ran for at most 150 epochs or until the validation loss stopped decreasing for a period of a least 10 epochs (or 6 epochs for CPC).
Dropout was applied to fully connected layers at a rate of 50\% and a weight decay of 0.001 was applied to the trainable parameters of all layers. 
Finally, the parameters of all neural networks were randomly initialized using uniform He initialization \cite{he2015delving}.

\subsection{Baselines}

The SSL tasks were compared to four baseline approaches on the downstream tasks: (1) random weights, (2) convolutional autoencoders, (3) purely supervised learning and (4) handcrafted features.

The random weights baseline used an embedder whose weights were frozen after random initialization.
The autoencoder (AE) was a more basic approach to representation learning, where a neural network made up of an encoder and a decoder learned an identity mapping between its input and its output, penalized by e.g., a mean squared error loss \cite{kramer1991nonlinear}.
Here, we used $h_\Theta$ as the encoder and designed a convolutional decoder that inverts the operations of $h_\Theta$.
The purely supervised model was directly trained on the downstream classification problem, i.e., it had access to the labeled data.
To do so, we added an additional linear classification layer to the embedder, before training the whole model with a multi-class cross-entropy loss.

Finally, we also included traditional machine learning baselines based on handcrafted features.
For sleep staging, we extracted the following features \cite{chambon2018deep}: mean, variance, skewness, kurtosis, standard deviation, frequency log-power bands between (0.5, 4.5, 8.5, 11.5, 15.5, 30) Hz as well as all their possible ratios, peak-to-peak amplitude, Hurst exponent, approximate entropy and Hjorth complexity.
This resulted in 37 features per EEG channel, which were concatenated into a single vector.
In the event of an artefact causing missing values in the feature vector of a window, we imputed missing values feature-wise using the mean of the feature computed over the training set.
For pathology detection, Riemannian geometry features were used like in \cite{gemein2020machine}, where it was reported that a nonlinear classifier trained on tangent space features reached high accuracy on the evaluation set of the TUH Abnormal dataset.
We did not average the covariance matrices per recording to allow a fair comparison with the other methods which work window-wise.
Therefore, for $C$ channels of EEG, the input to the classifier had dimensionality $C(C + 1)/2$.

For the downstream tasks, features learned with RP, TS, CPC and AE were classified using linear logistic regression with L2-regularization parameter\footnote{Varying $C$ had little impact on downstream performance, and therefore we used a value of 1 across experiments.} $C=1$, while handcrafted features were classified using a random forest classifier with 300 trees, maximum depth of 15 and a maximum number of features per split of $\sqrt{F}$ (where $F$ is the number of features)\footnote{Random forest hyperparameters were selected using a grid search with maximum depth in $\{3, 5, 7, 9, 11, 13, 15\}$, and maximum number of features per tree in $\{\sqrt{F}, \log_2{F}\}$ using the validation sets as described in Section~\ref{subsec:data}.}.
Balanced accuracy (bal acc), defined as the average per-class recall, was used to evaluate model performance on the downstream tasks.
Moreover, during training, the loss was weighted to account for class imbalance.
Models were trained using a combination of the braindecode \cite{schirrmeister2017deep}, MNE-Python \cite{gramfort2014mne}, pytorch \cite{paszke2017automatic},  pyRiemann \cite{BARACHANT2013172} and scikit-learn \cite{pedregosa2011scikit} packages.
Finally, deep learning models were trained on 1 or 2 Nvidia Tesla V100 GPUs for anywhere from a few minutes to 7h, depending on the amount of data, early stopping and GPU configuration. 


\subsection{Data}
\label{subsec:data}

The experiments were conducted on two publicly available EEG datasets, which are described in Tables~\ref{tab:pc18_desc} and \ref{tab:tuab_desc}.


\begin{table}[h]
\centering
\begin{tabular}{lc|lc}
\toprule
\multicolumn{4}{c}{{PC18 (train)}} \\
\midrule
& \# windows & & \\
W  & {158,020} & \# unique subjects & {994} \\
N1 & {136,858} & \# recordings & {994} \\
N2 & {377,426} & Sampling frequency & {200~Hz} \\
N3 & {102,492} & \# EEG channels & {6} \\
R & {116,872} & Reference & \multicolumn{1}{r}{{M1 or M2}} \\
Total & {891,668} &  & \\
\bottomrule
\end{tabular}
\caption{Description of the Physionet Challenge 2018 (PC18) dataset used in this study for sleep staging experiments.}
\label{tab:pc18_desc}
\end{table}

\begin{table}[h]
\centering
\begin{tabular}{lcc|lc}
\toprule
\multicolumn{5}{c}{{TUHab}} \\
\midrule
& train & eval & \# unique subjects & 2329 \\
& \# recordings & \# recordings & \# recordings & 2993 \\
Normal & 1371 & 150 & Sampling frequency & \multicolumn{1}{c}{250, 256, 512 Hz} \\
Abnormal & 1346 & 126 & \# EEG channels & \multicolumn{1}{c}{27 to 36} \\
Total & 2717 & 276 & Reference & Common average \\ 
\bottomrule
\end{tabular}
\caption{Description of the TUH Abnormal (TUHab) dataset used in this study for EEG pathology detection experiments.}
\label{tab:tuab_desc}
\end{table}

\subsubsection{Physionet Challenge 2018 dataset}

First, we conducted sleep staging experiments on the Physionet Challenge 2018 (PC18) dataset \cite{ghassemi2018you, goldberger2000physiobank}.
This dataset was initially released in the context of an open-source competition on the detection of arousals in sleep recordings, i.e., short moments of wakefulness during the night.
A total of 1,983 different individuals with (suspected) sleep apnea were monitored overnight and their EEG, EOG, chin EMG, respiration airflow and oxygen saturation measured.
Specifically, 6 EEG channels from the international 10/20 system were recorded at 200~Hz: F3-M2, F4-M1, C3-M2, C4-M1, O1-M2 and O2-M1.
The recorded data was then annotated by 7 trained scorers following the AASM manual \cite{berry2012aasm} into sleep stages (W, N1, N2, N3 and R).
Moreover, 9 different types of arousal and 4 types of sleep apnea events were identified in the recordings.
As the sleep stage annotations are only publicly available on about half the recordings (used as the training set during the competition), we focused our analysis on these 994 recordings.
In this subset of the data, mean age is 55 years old (min: 18, max: 93) and 33\% of participants are female.

\subsubsection{TUH Abnormal EEG dataset}

We used the TUH Abnormal EEG dataset v2.0.0 (TUHab) to conduct experiments on pathological EEG detection \cite{lopez2017automated}.
This dataset, a subset of \cite{obeid2016temple}, contains 2,993 recordings of 15 minutes or more from 2,329 different patients who underwent a clinical EEG in a hospital setting.
Each recording was labeled as ``normal'' (1,385 recordings) or ``abnormal'' (998 recordings) based on detailed physician reports.
Most recordings were sampled at 250~Hz (although some were sampled at 256 or 512~Hz) and contained between 27 and 36 electrodes.
Moreover, the corpus is divided into a training and an evaluation set with 2,130 and 253 recordings each.
The mean age across all recordings is 49.3 years old (min: 1, max: 96) and 53.5\% of recordings are of female patients.

\subsubsection{Data splits and sampling}

We split the available recordings from PC18 and TUHab into training, validation and testing sets such that the examples from each recording were only in one of the sets (see Table~\ref{tab:examples}).

For PC18, we used a 60-20-20\% random split, meaning there were 595, 199 and 199 recordings in the training, validation and testing sets respectively.
For RP and TS, 2,000 pairs or triplets of windows were sampled from each recording.
For CPC, the number of batches to extract from each recording was computed as 0.05 times the number of windows in that recording; moreover, we set the batch size to 32.

For TUHab, we used the provided evaluation set as the test set.
The recordings of the development set were split 80-20\% into a training and a validation set.
Therefore, we used 2,171, 543 and 276 recordings in the training, validation and testing sets.
Since the recordings were shorter for TUHab, we randomly sampled 400 RP pairs or TS triplets instead of 2000 from each recording.
We used the same CPC sampling parameters as for PC18.

\begin{table}[]
\centering
\begin{tabular}{lrrr|rrr}
\toprule
{\color[HTML]{000000} }      & \multicolumn{1}{l}{{\color[HTML]{000000} \textbf{PC18}}}        & \multicolumn{1}{l}{{\color[HTML]{000000} RP/TS}}     & \multicolumn{1}{l|}{{\color[HTML]{000000} CPC}}          & \multicolumn{1}{l}{{\color[HTML]{000000} \textbf{TUHab}}}       & \multicolumn{1}{l}{{\color[HTML]{000000} RP/TS}}     & {\color[HTML]{000000} CPC}          \\
{\color[HTML]{000000} }      & \multicolumn{1}{l}{{\color[HTML]{000000} \# recordings}} & \multicolumn{1}{l}{{\color[HTML]{000000} \# tuples}} & \multicolumn{1}{l|}{{\color[HTML]{000000} \# sequences}} & \multicolumn{1}{l}{{\color[HTML]{000000} \# recordings}} & \multicolumn{1}{l}{{\color[HTML]{000000} \# tuples}} & {\color[HTML]{000000} \# sequences} \\ \midrule
{\color[HTML]{000000} Train} & {\color[HTML]{000000} 595}                             & {\color[HTML]{000000} 1,190,000}                     & {\color[HTML]{000000} 877,792}                           & {\color[HTML]{000000} 2,171}                           & {\color[HTML]{000000} 868,400}                       & {\color[HTML]{000000} 642,144}            \\
{\color[HTML]{000000} Valid} & {\color[HTML]{000000} 199}                             & {\color[HTML]{000000} 398,000}                       & {\color[HTML]{000000} 294,272}                           & {\color[HTML]{000000} 543}                             & {\color[HTML]{000000} 217,200}                       & {\color[HTML]{000000} 160,224}            \\
{\color[HTML]{000000} Test}  & {\color[HTML]{000000} 199}                             & {\color[HTML]{000000} 398,000}                       & {\color[HTML]{000000} 292,608}                           & {\color[HTML]{000000} 276}                             & {\color[HTML]{000000} 110,400}                       & {\color[HTML]{000000} 81,184}            \\
\midrule
{\color[HTML]{000000} Total} & {\color[HTML]{000000} 993}                             & {\color[HTML]{000000} 1,986,000}                     & {\color[HTML]{000000} 1,464,672}                         & {\color[HTML]{000000} 2,990}                           & {\color[HTML]{000000} 1,196,000}                     & {\color[HTML]{000000} 883,552} \\
\bottomrule
\end{tabular}
\caption{Number of recordings used in the training, validation and testing sets with PC18 and TUHab, as well as the number of examples for each pretext task.}
\label{tab:examples}
\end{table}

\subsubsection{Preprocessing}

The preprocessing of the EEG recordings differed for the two datasets.
On PC18, the raw EEG was first filtered using a 30\,Hz FIR lowpass filter with a Hamming window, to reject higher frequencies that are not critical for sleep staging \cite{chambon2018deep,aboalayon2016sleep}.
The EEG channels were then downsampled to 100\,Hz to reduce the dimensionality of the input data.
For the same reason, we focused our analysis on channels F3-M2 and F4-M1.
Lastly, non-overlapping windows of 30\,s of size (3000 x 2) were extracted.

On TUHab, a similar procedure to the one reported in \cite{gemein2020machine} was used.
The first minute of each recording was cropped to remove noisy data that occurs at the beginning of recordings.
Longer files were also cropped such that a maximum of 20 minutes was used from each recording.
Then, 21 channels that are common to all recordings were selected (Fp1, Fp2, F7, F8, F3, Fz, F4, A1, T3, C3, Cz, C4, T4, A2, T5, P3, Pz, P4, T6, O1 and O2).
EEG channels were downsampled to 100\,Hz and clipped at $\pm 800\,\mu V$ to mitigate the effect of large artifactual deflections in the raw data.
Non-overlapping 6-s windows were extracted, yielding windows of size ($600 \times 21$).

Finally, windows from both datasets with peak-to-peak amplitude below 1\,$\mu V$ were rejected.
The remaining windows were normalized channel-wise to have zero-mean and unit standard deviation.

\section{Results}
\label{sec:experiments}

We investigated the use of SSL tasks to learn useful EEG features from unlabeled data in a series of three experiments.
First, SSL approaches were compared to fully supervised approaches based on deep learning or handcrafted features.
Second, we explored SSL-learned representations to highlight clinically-relevant structure.
Finally, in the last experiment, we studied the impact of hyperparameter selection on pretext and downstream performance.

\subsection{SSL models learn representations of EEG and facilitate downstream tasks with limited annotated data}
\label{subsec:data_quantity_exp}

Can the suggested pretext tasks enable SSL on clinical EEG data and mitigate the amount of labeled EEG data that is required in clinical tasks?
To address this question, we applied the pretext tasks to two clinical datasets (PC18 and TUHab) and compared their downstream performance to the one of various established approaches such as fully supervised learning, while varying the number of labeled examples available.

\textit{Context and setup.}
Feature extractors $h_\Theta$ were trained using the different approaches (AE, RP, TS and CPC on unlabeled data) and then used to extract features.
Following hyperparameter search (see Section~\ref{subsec:exp_hyperparameters}), we used same-recording negative sampling on PC18 and across-recording negative sampling on TUHab.
We also extracted features with randomly initialized models.
Downstream task performance was then evaluated by training linear logistic regression models on labeled examples, where the training set contains at least one and up to all existing labeled examples.
Additionally, fully supervised models were trained directly on labeled data and random forests were trained on handcrafted features.

\begin{figure}[t]
    \centering
    \includegraphics[width=\textwidth]{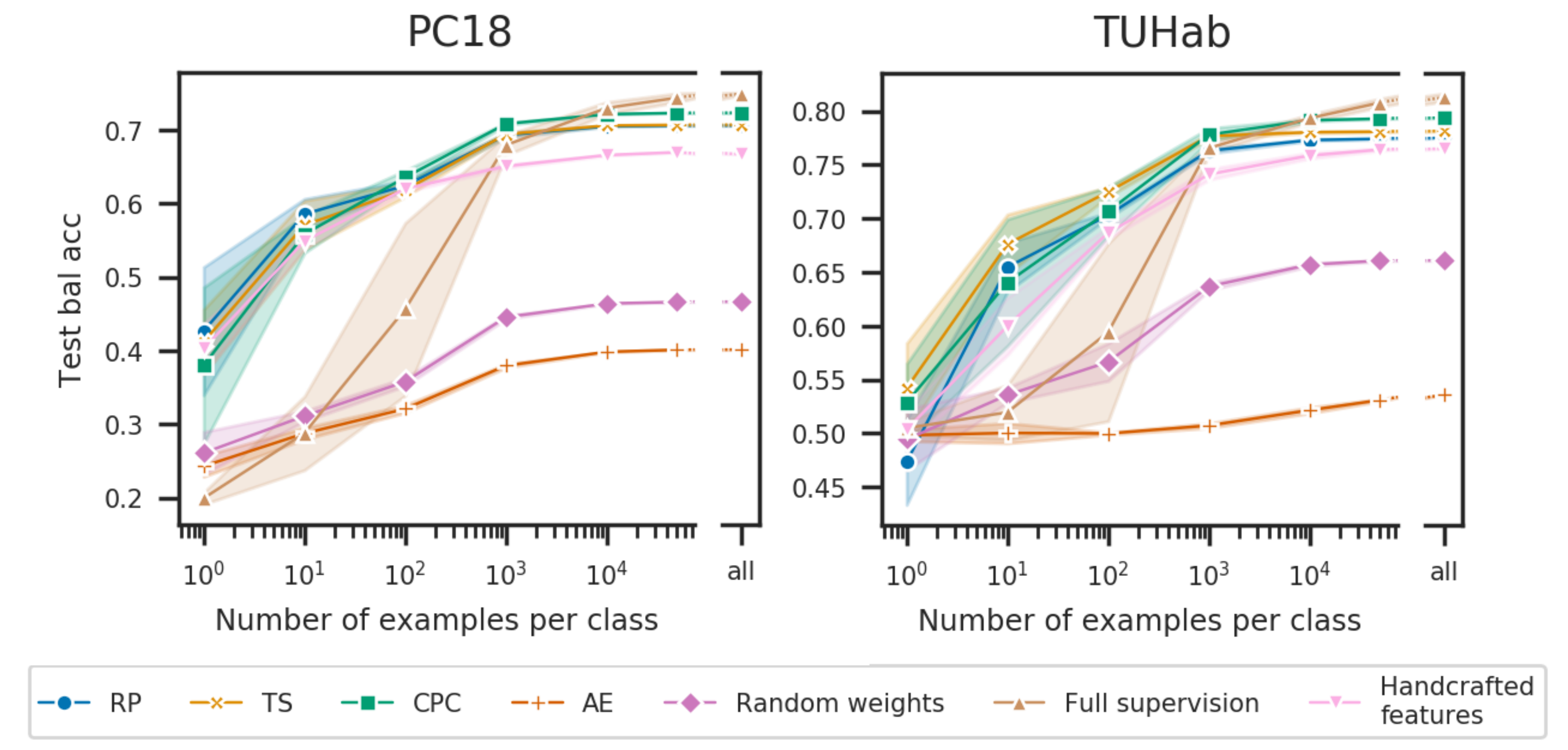}
    \caption{Impact of number of labeled examples per class on downstream performance. Feature extractors were trained with an autoencoder (AE), the relative positioning (RP), temporal shuffling (TS) and contrastive predictive coding (CPC) tasks, a fully supervised model, a model with random weights and handcrafted features, on PC18 and on TUHab. “All” means all available training examples were used.
    The same-recording negative sampling strategy was used for SSL tasks on PC18, while the across-recording strategy was used on TUHab.
    Results are the average of five runs with same initialization but different random selection of examples and shaded area represents standard deviation.
    While higher numbers of labeled examples led to better performance, SSL models achieved much higher performance than a fully supervised model when few labeled examples were available.}
    \label{fig:data_quantity_exp}
\end{figure}

The impact of the number of labeled samples on downstream performance is presented in Fig.~\ref{fig:data_quantity_exp}.
First, when using SSL-learned features for the downstream tasks, we observe important above-chance performance across all data regimes: on PC18, our models scored as high as 72.3\% balanced accuracy (5-class, chance=$20\%$) while on TUHab the highest performance was of 79.4\% (2-class, chance=$50\%$).
These results demonstrate the ability of SSL to learn useful representations for our downstream tasks.
Second, the comparison suggests that SSL-learned features are competitive with other baseline approaches and can even outperform supervised approaches.
On the PC18 sleep data (Fig.~\ref{fig:data_quantity_exp}\textbf{A}), one can observe that all three SSL pretext tasks outperformed alternative approaches including the fully supervised model and handcrafted features in most data regimes.
The performance gap between SSL-learned features and full supervision was as high as 22.8 points when only one example per class was available. 
It remained in favor of SSL up to around 10,000 examples per class, where full supervision finally began to exceed SSL performance, however by a 1.6-3.5\% margin only.
Moreover, SSL outperformed the handcrafted features baseline over 100 examples per class, e.g., by up to 5.6 points for CPC.
These results suggest two important implications: (1) our pretext tasks can capture critical information for sleep staging, even though no sleep labels were available when learning representations and (2) these features can rival both human-engineered features and label-intensive full supervision.

Other baselines such as random weights and autoencoding obtained much lower performance, showing that learning informative features for sleep staging is not trivial and requires more sophistication than the inductive bias of a convolutional neural network alone or a pure reconstruction task. 
Interestingly, the poor performance of the AE can be attributed to its mean squared error loss. 
This encourages the model to focus on the signal's low frequencies, which, due to 1/f power-law dynamics have the largest amplitudes in biosignals like EEG. 
Yet, low frequency signals only capture a small portion of the neurobiological information in EEG signals.

Next, we applied SSL to the task of pathology detection, where the two classes (``normal'' and ``abnormal'') are likely to be more heterogenous than the sleep staging classes.
Again, SSL-learned features outperformed the baseline approaches in most data regimes: CPC outperformed full supervision when fewer than 10,000 labeled examples per class were available, while the performance gap between the two methods was on the order of 1\% when all examples were available.
Handcrafted features were also consistently outperformed by RP, TS and CPC, albeit by a smaller amount (e.g., 3.8-4.8 point difference for CPC).
Again, the AE and random weights features could not compete with the other methods.
Notably, the AE fared even worse on TUHab and downstream performance never exceeded 53.0\%.

Taken together, our results demonstrate that the proposed SSL pretext tasks were general enough to enable two fundamentally different types of EEG classification problems.
All SSL tasks systematically outperformed or equaled other approaches in low-to-medium labeled data regimes and remained competitive in a high labeled data regime.

\begin{figure}[!t]
    \centering
    \includegraphics[width=\textwidth]{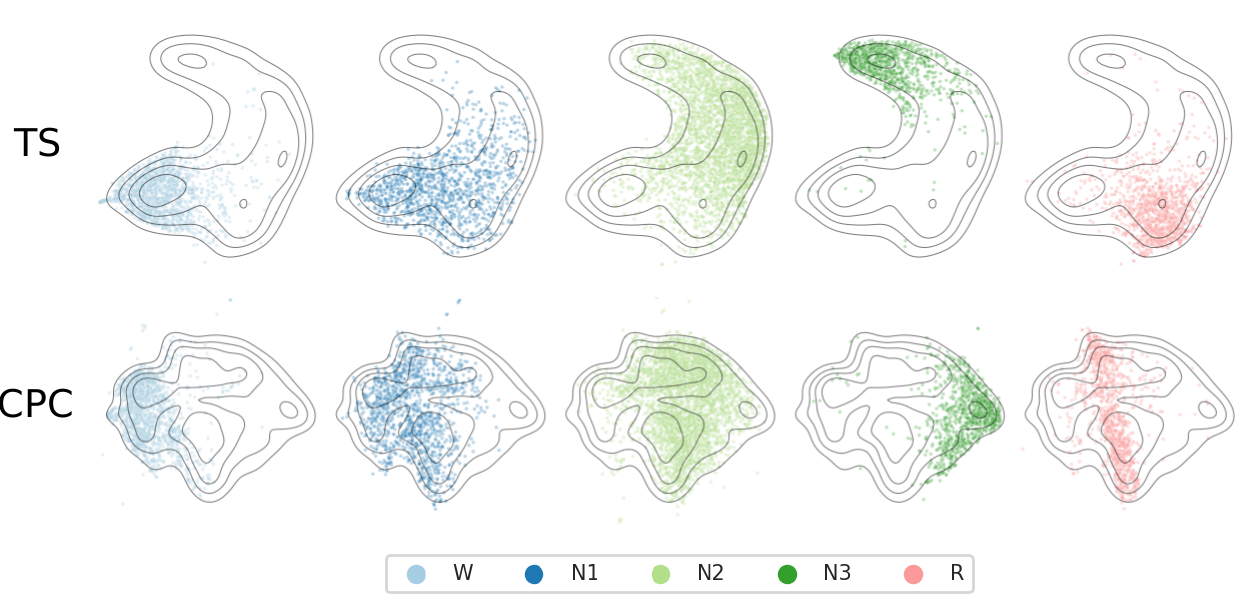}
    \caption{UMAP visualization of SSL features on the PC18 dataset. The subplots show the distribution of the 5 sleep stages as scatterplots for TS (first row) and CPC (second row) features. Contour lines correspond to the density levels of the distribution across all stages and are used as visual reference. Finally, each point corresponds to the features extracted from a 30-s window of EEG.
    In both cases, there is clear structure related to sleep stages although no labels were available during training.}
    \label{fig:umaps_stages}
\end{figure}

\begin{figure}[!t]
    \centering
    \includegraphics[width=\textwidth]{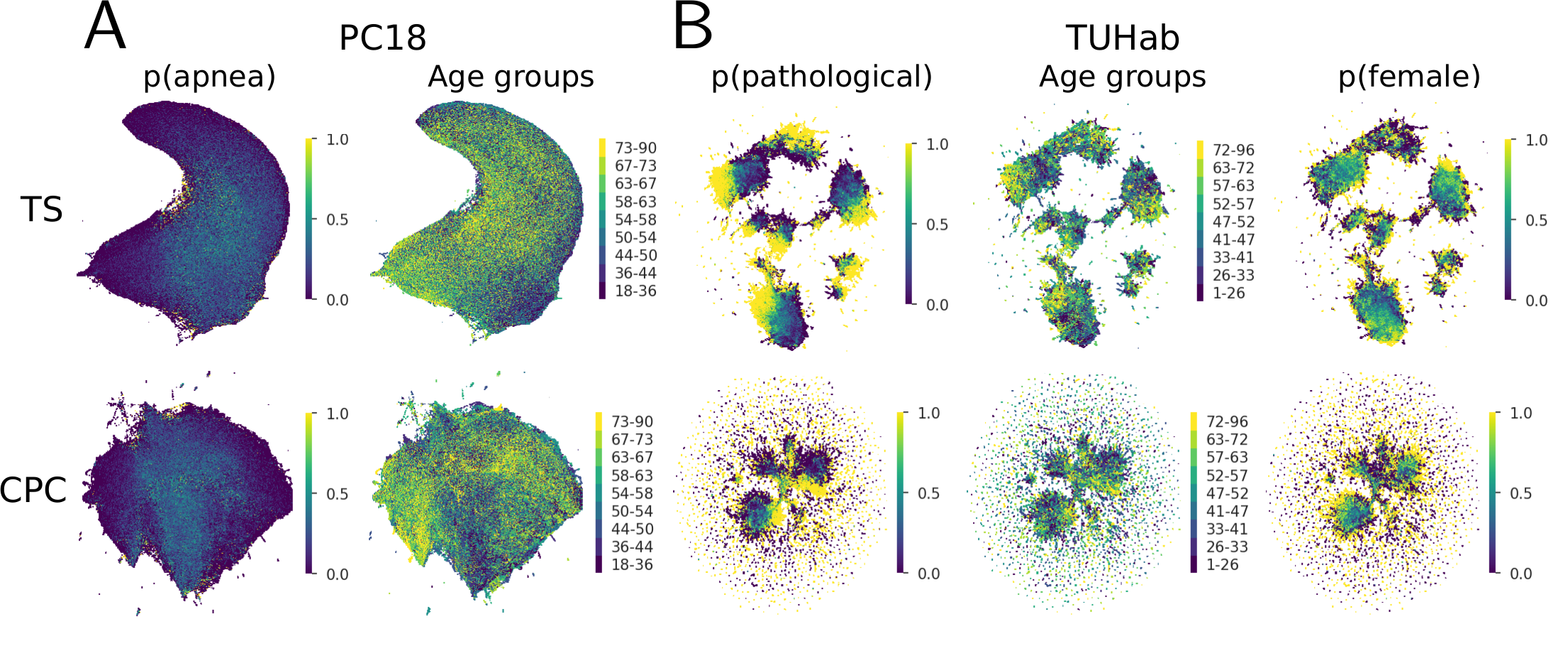}
    \caption{Structure learned by the embedders trained on SSL tasks.
    The 2D embedding spaces obtained with UMAP on (\textbf{A}) PC18 and (\textbf{B}) TUHab were discretized into 500 x 500 ``pixels''.
    For binary labels (``apnea'', ``pathological'' and ``gender''), we visualize the probability as heatmaps, i.e., the color indicates the probability that the label is true (e.g., that a window in that region of the embedding overlaps with an apnea annotation).
    For age, the subjects of each dataset were divided into 9 quantiles, and the color indicates which group was the most frequent in each bin.
    The features learned with the SSL tasks capture physiologically-relevant structure, such as pathology, age, apnea and gender.}
    \label{fig:umaps_others}
\end{figure}

\begin{figure}
    \centering
    \includegraphics[width=\textwidth]{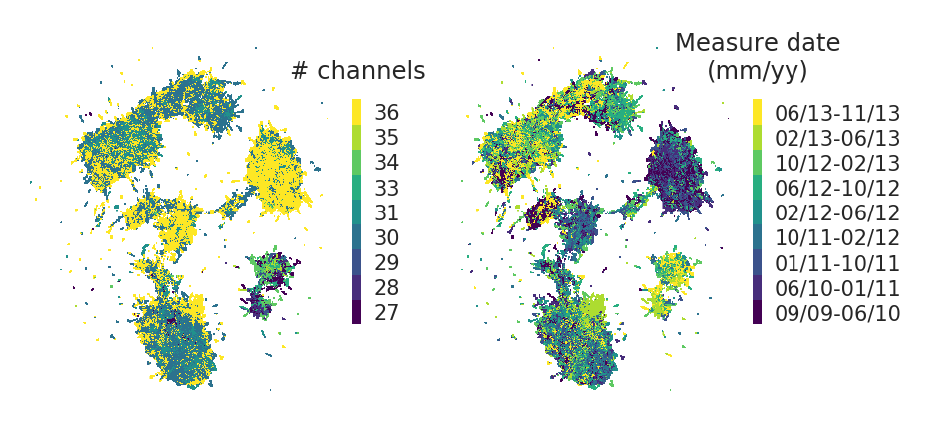}
    \caption{Structure related to the original recording's number of EEG channels and measurement date in TS-learned features on the entire TUHab dataset. 
    The overall different number of EEG channels and measurement date in each cluster shows that the cluster-like structure reflects differences in experimental setups. 
    See Fig.~\ref{fig:umaps_others} for a description of how the density plots are computed.}
    \label{fig:umaps_tuh_clusters}
\end{figure}

\subsection{SSL models capture physiologically and clinically meaningful features}
\label{subsec:umap_exp}

While SSL-learned features yielded competitive performance on sleep staging and pathology detection tasks, it is unclear what kind of structure was captured by SSL.
To address this, we examined the embeddings by analyzing their relationship with the different annotations and meta-data available in clinical datasets.
We thus projected the 100-dimensional embeddings obtained on PC18 and TUHab onto a two-dimensional representation using Uniform Manifold Approximation and Projection (UMAP) \cite{mcinnes2018umap} and using the best models as identified in Section~\ref{subsec:exp_hyperparameters}.
This allows a qualitative analysis of the local and global structure of the SSL-learned features.

Results on sleep data are shown in Fig.~\ref{fig:umaps_stages}.
A structure that closely follows the different sleep stages can be noticed in the embeddings of PC18 obtained with TS and CPC\footnote{Results for RP were similar to TS's and are presented in Appendix.}.
Upon inspection of the distribution of examples from the different stages, clear groups emerged. 
They not only corresponded to the labeled sleep stages, but they are also sequentially arranged: moving from one end of the embedding to another, we can draw a trajectory that passes through W, N1, N2 and N3 sequentially.
Stage R, finally, mostly overlaps with N1.
These results are in line with previous observations on the structure of the sleep-wakefulness continuum \cite{pardey1996new,lopour2011continuous}.

Intuitively, the largest sources of variation in sleep EEG data are linked to changes in sleep stages and the corresponding microstructure (e.g., slow waves, sleep spindles, etc.).
Can we expect other sources of variation to also be visible in the embeddings?
To address this question, we inspected clinical information available in PC18 along with the embeddings: apnea events and subject age.
The results are presented in Fig.~\ref{fig:umaps_others}.
First, apnea-related structure can be seen in the middle of the embeddings, overlapping with the area where stage N2 was prevalent (first column of Fig.~\ref{fig:umaps_others}).
At the same time, very few apnea events occurred at the extremities of the embedding, for instance over W regions, naturally, but also over N3 regions.
Although this structure likely reflects the correlation between sleep stages, age and actual apnea-induced EEG patterns, this nonetheless shows the potential of SSL to learn features that relate to clinical phenomena.
Second, age structure was revelead in at least two distinct ways in the embeddings (second column of Fig.~\ref{fig:umaps_others}).
The first is related to sleep macrostructure, i.e., the sequence of sleep stages and their relative length.
Indeed, younger subjects were predominant over the R stage region, while older subjects were more frequently found over the W region.
This is in line with well-documented phenomena such as increased sleep fragmentation and sleep onset latency in older individuals, as well as a subtle reduction in REM sleep with age \cite{mander2017sleep}.
Concurrently, we also observe sleep microstructure in the embeddings.
For instance, looking at N2-N3 regions for the TS embedding, older age groups are more likely to be found in the leftmost side of the blob, while younger subjects are more likely to be found on its rightmost side.
This suggests there is a difference between the characteristics of N2-N3 sleep across age groups, e.g., related to sleep spindles \cite{purcell2017characterizing}.

Can this clinically-relevant structure also be learned on a different type of EEG recording?
We conducted a similar analysis for TUHab, this time focusing on pathology, age and gender.
Results are shown in columns 3-5 of Fig.~\ref{fig:umaps_others}B.
The embeddings of both TS and CPC exhibited a primary multi-cluster structure, with similar gradient-like structure inside each cluster\footnote{Results for RP are presented in Appendix.}. 
For instance, pathology-related structure was clearly exposed in the two embeddings (column 3), with an increasing probability of the EEG being abnormal when moving from one end of the different clusters to the other.
Likewise, an age-related gradient emerged inside each cluster (column 4), in a similar direction as the pathology gradient, while there is also a gender-associated gradient that appeared orthogonal to the first two (last column).
What do the different clusters actually represent?
We plotted experimental setup-related labels (the original number of EEG channels and the measurement date of each recording) in Fig.~\ref{fig:umaps_tuh_clusters}.
Each cluster was predominantly composed of examples with a given number of channels and with a specific range of measurement dates.
This might suggest that the SSL tasks have partially learned the noise introduced by data collection.
For example, the TUHab dataset was collected over many years across different sites, by different EEG technicians and with various EEG devices.
Most likely, the impact of this noise in the embedding could be mitigated by using more aggressive preprocessing (e.g., bandpass filtering) or by sampling negative examples within recordings from the same cohort.

In conclusion, this experiment showed that SSL can encode clinically-relevant structure such as sleep stages, pathology, age, apnea and gender information from EEG data, while revealing interactions (such as young age and REM sleep), without any access to labels.

\subsection{SSL pretext task hyperparameters strongly influence downstream task performance}
\label{subsec:exp_hyperparameters}

How should the various SSL pretext task hyperparameters be tuned to fully make use of self-supervision in clinical EEG tasks?
In this section, we describe how the hyperparameters of the models used in the experiments above were tuned and study in detail the impact of some key hyperparameters on downstream performance.

\begin{figure}[t]
    \centering
    \includegraphics[width=\textwidth]{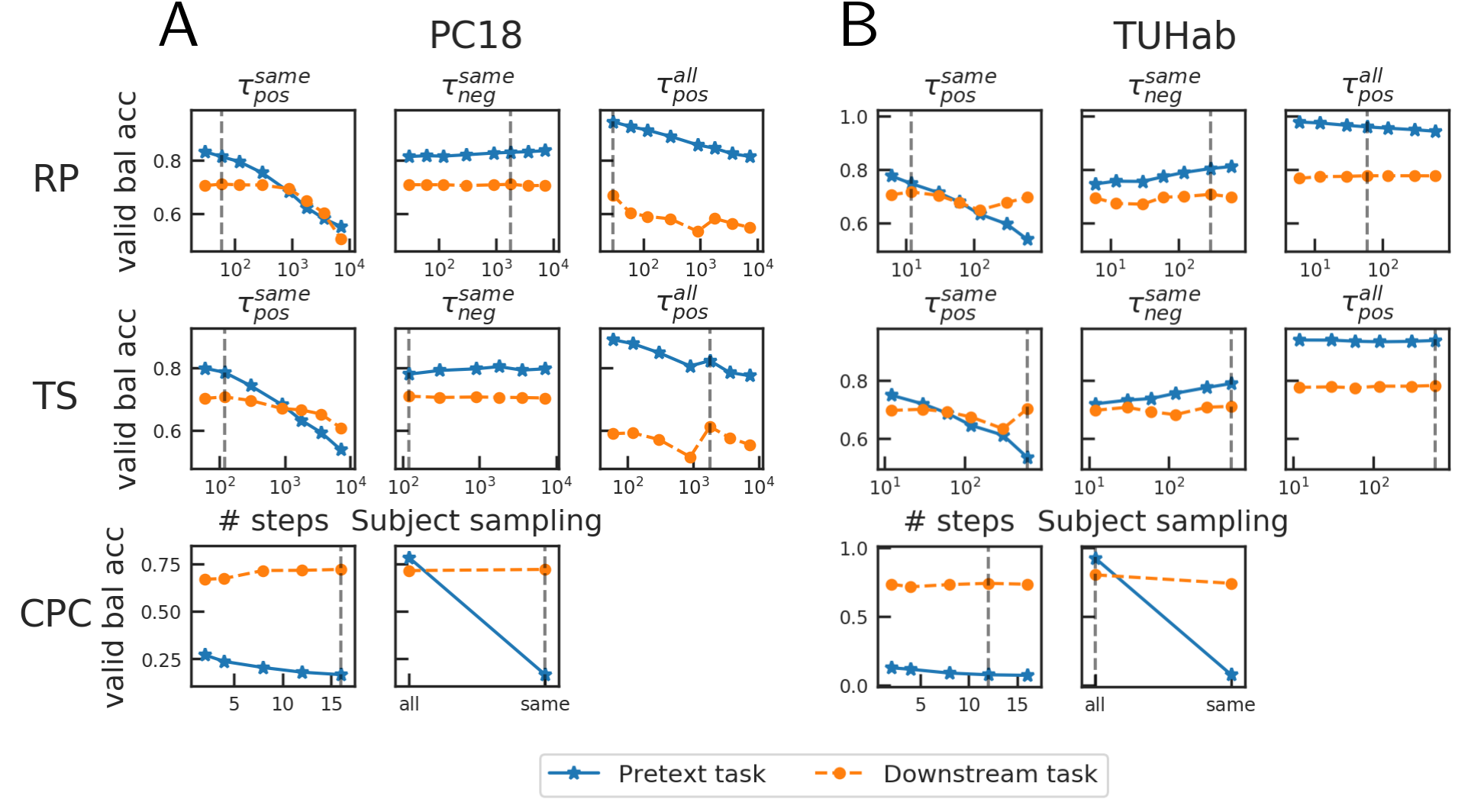}
    \caption{Impact of principal hyperparameters on pretext (blue) and downstream (orange) task performance, measured with balanced accuracy on the validation set on (A) PC18 and (B) TUHab. 
    Each row corresponds to a different SSL pretext task.
    For both RP and TS, we varied the hyperparameters that control the length of the positive and negative contexts ($\tau_{pos}$, $\tau_{neg}$, in seconds); the exponent ``same'' or ``all'' indicates whether negative windows were sampled across the same recording or across all recordings, respectively.
    For CPC, we varied the number of predicted windows and the type of negative sampling.
    Finally, the best hyperparameter values in terms of downstream task performance are emphasized using vertical dashed lines.
    See text for more details on the hyperparameter search procedure.}
    \label{fig:hyperparams}
\end{figure}

To benchmark different pretext tasks across datasets, we tracked the performance of the pretext and downstream tasks across different choices of hyperparameters (see Section~\ref{subsec:hp_search_procedure} for a complete description of the search procedure).
The comparison is depicted in Fig.~\ref{fig:hyperparams}.
The analysis suggests that the pretext tasks performed significantly above chance level on all datasets: RP and TS reached a maximum performance of 98.0\% (2-class, chance=$50\%$) while CPC yielded performances as high as 95.4\% (32-class, chance$=3.1\%$).
On the downstream tasks, SSL-learned representations always performed above chance as reported in Section~\ref{subsec:data_quantity_exp}.
Interestingly though, configurations with high pretext performance did not necessarily lead to high downstream performance, which highlights the necessity of appropriate hyperparameter selection.

In the next step, we examined the influence of the different hyperparameters on each pretext task (rows of Fig.~\ref{fig:hyperparams}) to identify optimal configurations.
First, we focused on the same-recording negative sampling scenario, in which negative examples are sampled from the same recording as the anchor window(s).
With RP, increasing $\tau_{pos}$ always made the pretext task harder.
This is expected since the larger the positive context, the more probable it is to get positive example pairs that are composed of distant (and thus potentially dissimilar) windows.
On sleep data, we notice a plateau effect: the downstream performance was more or less constant below $\tau_{pos}=20$ min, suggesting EEG autocorrelation properties might be changing at this temporal scale.
Although this phenomenon did not appear clearly on TUHab, downstream performance decreased above $\tau_{pos}=30$ s, and then increased again after $\tau_{pos}=2$ min.
On the other hand, varying $\tau_{neg}$ given a fixed $\tau_{pos}$ did not have a consistent or significant influence on downstream performance, although larger $\tau_{neg}$ generally led to easier pretext tasks.

Do these results hold when negative windows are sampled across all recordings?
Interestingly, the type of negative sampling has a considerable effect on downstream performance (columns 3 and 6 of Fig.~\ref{fig:hyperparams}).
On sleep staging, downstream performance dropped significantly and degraded faster as $\tau_{pos}$ was increased, while the opposite effect could be seen on the pathology detection task (higher, more stable performance).
This effect might be explained by the nature of the downstream task: in sleep staging, major changes in the EEG occur inside a given recording, therefore distinguishing between windows of a same recording is key to identifying sleep stages.
On the other hand, in pathology detection, each recording is given a single label (``normal'' or ``pathological'') and so being able to know whether a window comes from the same recording (necessarily with the same label) or from another one (possibly with the opposite label) intuitively appears more useful.
In other words, the distribution that is chosen for sampling negative examples determines the kind of invariance that the network is forced to learn.
Overall, similar results hold for TS. 

As for CPC, a similar analysis shows that while increasing the number of windows to predict (``number of steps'') made the pretext task more difficult, predicting further ahead in the future helped the embedder learn a better representation for sleep staging (bottom row of Fig.~\ref{fig:hyperparams}).
Were these results affected by the type of negative sampling as was the case for RP and TS?
Remarkably, the type of negative sampling had a minor effect on downstream performance on sleep data ($71.6$ vs. $72.2\%$ bal acc), but had a considerable effect on pathology detection ($74.1$ vs. $80.4\%$), akin to what was seen above on RP and TS.
This result echoes the report of \cite{oord2018representation} where subject-specific negative sampling led to the highest downstream performance on a phoneme classification downstream task.

In this last experiment, we confirmed that our SSL pretext tasks are not trivial, and that certain pretext task hyperparameters have a measurable impact on downstream performance.

\section{Discussion}
\label{sec:discussion}

In this paper, we introduced self-supervised learning (SSL) as a way to learn representations on EEG data.
Specifically, we proposed two SSL tasks designed to capture structure in EEG data, relative positioning (RP) and temporal shuffling (TS) and adapted a third approach, contrastive predictive coding (CPC), to work on EEG data.
We showed that these tasks can be used to learn generic features which capture clinically relevant structure from unlabeled EEG, such as sleep micro- and macrostructure and pathology.
Moreover, we performed a rigorous comparison of SSL methods to traditional unsupervised and supervised methods on EEG, and showed that downstream classification performance can be significantly improved by using SSL, particularly in low-labeled data regimes.
These results hold for two large-scale EEG datasets comprising sleep and pathological EEG data, both with thousands of recordings.

\subsection{Using SSL to improve performance in semi-supervised scenarios}
\label{subsec:improving_perf}

We showed that SSL can be used to improve downstream performance when a lot of unlabeled data is available but the amount of labeled examples is limited, i.e., in a semi-supervised learning scenario (see Section~\ref{subsec:improving_perf}).
For instance, CPC-learned features outperformed fully supervised learning on sleep data by about 20\% when only one labeled example per class was available.
Similarly, on the pathology detection task an improvement of close to 15\% was obtained with SSL when only 10 labeled examples per class were available.
In practice, SSL has the potential to become a common tool for boosting classification performance when annotations are expensive, a common scenario when working with biosignals like EEG.

Can SSL be applied to any semi-supervised problem or is it limited to larger clinical datasets?
While the SSL pretext tasks included in this work are applicable to multivariate time series in general, their successful application does require adequate recording length and dataset size.
First, the EEG recordings need to be sufficiently long so that a reasonable number of windows can be sampled given the positive and negative contexts and the window length.
For clinical recordings, this is typically not an issue: sleep monitoring and pathology screening procedures both produce recordings of tens of minutes to many hours, which largely suffice.
It is noteworthy that while the proposed SSL approaches have been developed using clinical data, they could be readily applied to event-related EEG protocols, such as those encountered in cognitive psychology or brain-computer interfacing experiments \cite{woodman2010brief,abiri2019comprehensive}.
Indeed, SSL could be applied to an entire recording without explicitly taking into consideration the known events (e.g. stimuli, behavioral responses).
This critically depends on the availability of continuous recordings (rather than epoched data only, as is the case in some public EEG datasets) including resting state baselines and between-trial segments.
Second, the current results may suggest large datasets are necessary to enable SSL on EEG as analyses were based on two of the largest publicly available EEG datasets with thousands of recordings each.
However, similar results hold on much smaller datasets containing fewer than 100 recordings \cite{banville2019self}.
Intuitively, as long as the unlabeled dataset is representative of the variability of the test data, representations that are useful for the pretext task should be transferable to a related downstream task.

One might argue that the observed performance benefit of SSL is incremental as compared to supervised approaches when the number of samples is moderate to large. 
Is investing in the development of SSL-based EEG-analysis worth the effort?
It is important to highlight that our results present a proof of concept that opens the door to further developments, which may lead to substantial improvements in performance. 
In this paper, we limited our experiments to the linear evaluation protocol, where the downstream task is carried out by a linear classifier trained on SSL-learned features, in order to focus on the properties of the learned representations.
Finetuning the parameters of the embedders on the downstream task \cite{oord2018representation,tian2020makes} could likely further improve downstream performance.
Preliminary experiments (not shown) suggested that a 3 to 4-point improvement can be obtained in some data regimes when finetuning the embedders and that performance with all data points is as high as with a purely supervised model.
However, using nonlinear classifiers (here, random forests) as suggested in \cite{oord2018representation} on SSL-learned features did not improve results, suggesting our downstream tasks might be sufficiently close to the pretext task that the relevant information is already linearly accessible.
Another potential opportunity to improve downstream performance consists of using larger deep learning architectures.
Due to their sampling methodology, most SSL tasks can ``create'' a vast number of distinct examples going far beyond the number of labeled examples typically available in a supervised task.
For instance, on PC18, our training set contained close to $5 \times 10^5$ labeled examples while for our self-supervised tasks we chose to train our embedders on more than twice that number of examples (to fit computational time requirements), though more could have been easily sampled.
The much higher number of available examples in SSL opens the door to using much larger deep neural networks which require much larger sample sizes ~\cite{henaff2019data}.
Given the relatively shallow architectures currently used in deep learning and EEG research (on which we based our choice of architectures) \cite{roy2019deep}, SSL could be key to training deeper models and improving upon current state of the art on various EEG tasks.

\subsection{Sleep-wakefulness continuum and inter-rater reliability}

We have demonstrated that the embeddings learned with SSL capture clinically-relevant information.
Sleep, pathology, age and gender information appeared as clear structure in the learned feature space (see Fig.~\ref{fig:umaps_stages}-\ref{fig:umaps_others}).
The variety of metadata that is visible in the embeddings highlights the capacity of the proposed SSL tasks to uncover important factors of variation in noisy biosignal data such as EEG in a purely unsupervised manner.
Critically though, this structure is not discrete, but continuous.
Indeed, sleep stages are not cleanly separated into five clusters (or two clusters for normal and abnormal EEG), but instead the embeddings display a smooth continuum of sleep-wakefulness (or of normal-abnormal EEG).
Is this gradient-like structure meaningful, or is it a mere artefact of our experimental setup?
We argue that the continuous nature of SSL-learned features is inherent to the neurophysiological phenomena under study.
Conveniently, this offers interesting opportunities to improve the analysis of physiological data.

While sleep EEG is routinely divided into discrete stages to be analyzed, its true nature is likely continuous. 
For instance, the concept of sleep stages and the taxonomy that we know today is the product of incremental standardization efforts in the sleep research community \cite{loomis1937cerebral,kales1968manual,moser2009sleep,berry2012aasm}.
Although this categorization is convenient, critics still stress the limitations of stage-based systems under the evidence of sub-stages of sleep and the interplay between micro- and macrostructure \cite{schulz2008rethinking}.
Moreover, even trained experts using the same set of rules do not perfectly agree on their predictions, showing that the definition of stages remains ambiguous: in \cite{younes2016staging}, an overall agreement of 82.6\% was obtained between the predictions of more than 2,500 trained sleep scorers (even lower for N1, at 63.0\%).
Consequently, could a better representation of sleep EEG, such as the one learned with self-supervision, alleviate some of these challenges?
While previous research suggests sleep might indeed be measured using a continuous metric derived from a supervised machine learning model \cite{pardey1996new} or a computational mean-field model \cite{lopour2011continuous}, we additionally demonstrated that the rich feature space learned with SSL can simultaneously capture sleep-related structure and variability caused by age and apnea.
Importantly, the data-driven nature of the SSL representation alleviates the subjectivity of manual sleep staging.
Overall, this suggests SSL-learned representations could provide more fine-grained information about the multiple factors at play during sleep and, in doing so, enable a more precise study of sleep.

Similarly, many EEG pathologies are described by a clinical classification system which defines discrete subtypes of diseases or disorders, e.g., epilepsy \cite{pack2019epilepsy,england2012epilepsy} and dementia \cite{walters2010dementia}.
As for sleep EEG, inter-rater agreement is limited \cite{gemein2020machine}.
This suggests that there is an opportunity for these pathologies to be interpreted as a continuum as well. 
Although our pathology detection downstream task was a simple binary classification task, the clear pathology-associated gradient captured by our SSL pretext tasks could be used to characterize the different types and subtypes of pathologies contained in the dataset more precisely (see Fig.~\ref{fig:umaps_others}).
Ultimately, the data-driven feature space obtained with SSL might aid in the study and comparison of different neuropathologies and EEG patterns in general.


\subsection{Finding the right pretext task for EEG}
\label{subsec:right_pretext_task}

With the large number of self-supervised pretext tasks one can think of, and the even larger number of possible EEG downstream tasks, how can we choose a combination of pretext task and hyperparameters for a given setting?
To answer this question, many more empirical experiments will have to be conducted on EEG data.
However, the results presented here give some insight as to what may be important to consider.
In this work, we developed pretext tasks that proved effective on two different classification problems by using a combination of (1) prior knowledge about EEG time series, (2) assumptions about the statistical structure of the features to be learned, (3) thorough hyperparameter search and (4) computational considerations.

First, we introduced and tailored to EEG the relative positioning (RP) and temporal shuffling (TS) tasks by relying on prior knowledge about EEG.
These tasks were designed specifically with the structure of sleep data in mind.
Indeed, sleep EEG signals have a clear temporal structure originating from the succession of sleep stages during the night.
This means that two windows that are close in time have a high probability of sharing the same sleep stage annotation and statistical structure.
Therefore, learning to differentiate close-by from faraway windows should intuitively be related to learning to differentiate sleep stages.
Similar approaches were previously described in the computer vision literature \cite{doersch2015unsupervised,misra2016shuffle}, however they rely on properties of natural images that generally do not hold for EEG.
For instance, whereas two EEG windows $x_t$ and $x_{t'}$ that are close in time likely look alike, there is typically no physiological information in these windows that would allow one to determine whether $t < t'$ or $t > t'$.\footnote{Exceptions would include transitions between sleep stages that are more likely than others, such as from lighter to deeper sleep stages; however these transitions occur rarely during the night; more often a back-and-forth between sleep stages is observed.}
Therefore, tasks that rely on proximity rather than absolute positioning appear to be a better match for EEG.
As for CPC, we included it in our experiments as it is a natural extension of RP and TS and has led to promising results on other kinds of data \cite{oord2018representation}.

Second, assumptions about the statistical structure of the latent factors or features to recover was used to further support our choice of tasks.
For instance, given its similarity with permutation contrastive learning (PCL, a self-supervised method for performing nonlinear ICA \cite{hyva17aistats}), RP likely relies on the general temporal dependencies (including autocorrelations) of EEG signals to recover informative features.\footnote{PCL can be obtained by setting RP's $\tau_{pos}$ to be the length of a single window and $\tau_{neg}$ to 0. Incidentally, we found that the optimal value of $\tau_{pos}$ and $\tau_{neg}$ were relatively small on the datasets considered, suggesting hyperparameters close to those of PCL are optimal.}
Since TS and CPC can both be seen as extensions of the RP task with more elaborate sampling strategies and contrastive procedures (see Section~\ref{subsec:ssl}), all three tasks might effectively rely on similar structure to discover features.

Third, the careful selection of pretext task hyperparameters was essential to selecting the right pretext task configuration.
For instance, RP, TS and CPC often yielded very similar downstream task performance once the best hyperparameters were selected.
Out of the main pretext task hyperparameters, the negative sampling strategy proved to be especially important to tune (Section~\ref{subsec:exp_hyperparameters}).
Indeed, sleep staging benefited from same-recording negative sampling whereas pathology detection instead worked better when across-recording negative sampling was used.
Interestingly, this appears to be the unique change to RP, TS and CPC that is required to compete with purely supervised approaches on the pathology detection downstream task, although RP and TS were initially designed for capturing intra-recording sleep structure.
Thus, negative sampling hyperparameters might be among the most critical hyperparameters to tune, as they can be used to develop invariances to particular structure that is not desirable, e.g., intra-recording changes or measurement-site effects (Fig.~\ref{fig:umaps_tuh_clusters}). 
Ultimately, the fact that all three pretext tasks could reach similar downstream performance suggests self-supervision was able to uncover fundamental information likely related to physiology.

Finally, computational requirements and architecture-specific constraints are important to consider when choosing a pretext task. 
For example, being simpler, RP and TS might be preferred over CPC if they can reach similar performance.
Indeed, CPC has more hyperparameters than RP and TS (i.e., number of context windows, of windows to be predicted and of negative examples; architecture of the autoregressive embedder $g_{AR}$) and requires additional forward and backward passes through the embedder $h_{\Theta}$ at training time.
However, CPC's autoregressive encoder $g_{AR}$ could yield better features for some tasks with larger-scale dependencies \cite{oord2018representation}, e.g., sleep staging: indeed, previous studies on deep learning for automated polysomnography have reported improved performance using multiple consecutive windows \cite{chambon2018deep} or a recurrent network on top of a window encoder \cite{supratak2017deepsleepnet}.
This aggregation of window-level information is already part of a CPC model and therefore the pretrained $g_{AR}$ could be reused directly.
Preliminary results (not shown) suggest these autoregressive features can substantially improve downstream performance on both the sleep staging and pathology detection tasks.

Although the proposed tasks proved successful, many other pretext tasks could have been designed based on an intuitive understanding of EEG.
For instance, a transformation discrimination task was applied to ECG data in \cite{sarkar2020self} and could be adapted for EEG.
Similarly, nonlinear ICA-derived frameworks such as time contrastive learning \cite{hyvarinen2016unsupervised} and generalized contrastive learning \cite{hyvarinen2019nonlinear} could be used to explicitly leverage nonstationnarity or other structure present in EEG signals.




\subsection{Limitations}

We identify three principal limitations to this work: fixed training hyperparameters across data regimes, restricted architecture search, and difference between reported results and state of the art.

Deep neural networks can easily overfit the training data when few examples are available, which can negatively impact generalization.
Typical ways of addressing overfitting include increasing the strength of regularization such as dropout and weight decay and performing early stopping.
Given the computational requirements of training neural networks on large EEG datasets, we fixed the training hyperparameters of the fully supervised models (i.e., learning rate, batch size, dropout, weight decay) and reused the same values across all data regimes.
As a result, the fully supervised models typically stopped learning after only a few epochs, although they might have been able to train longer with different hyperparameters. 
We tested the impact of various training hyperparameter settings on a subset of the models and saw that even though training can be slightly improved by changing hyperparameters, this effect is not strong enough to change any of our conclusions (results not shown).

Similarly, hyperparameter search was limited to pretext task hyperparameters in our experiments.
However, architecture hyperparameters (e.g., number of convolutional channels, dimensionality of the embedding, number of layers, etc.) can also play a critical role in achieving high performance using SSL \cite{kolesnikov2019revisiting}.
Sticking to a single fixed architecture for all models and data regimes means that these improvements - which could help bridge (or widen) the gap between SSL methods and the various baselines - were not taken into account in this work.

Finally, the goal of this work being to introduce self-supervision as a representation learning paradigm for EEG, we did not focus on matching state-of-the-art performance from earlier work on sleep staging and pathology detection.
To the best of our knowledge, we are the first to present sleep staging results on PC18, and therefore there is no prior art to which our results can be fairly compared.
Nonetheless, downstream performance would most likely improve significantly by aggregating temporal windows as was shown on other datasets \cite{chambon2018deep,supratak2017deepsleepnet}.
Similarly, on TUHab, we chose to reuse simpler approaches from \cite{gemein2020machine} instead of the best performing models. 
Moreover, (1) we did not use the cropped decoding approach presented in \cite{schirrmeister2017deep}, (2) we used a simpler normalization methodology (z-score instead of exponential moving average normalization) and (3) our train/validation data split was different.
Together, these differences explain the small drop in performance between these state-of-the-art methods and the results reported here.

\section{Conclusion}



In this work, we introduced SSL approaches to learn representations on EEG data and showed that they could compete with and sometimes outperform traditional supervised approaches on two large clinical tasks.
Importantly, the features learned by SSL displayed a clear structure in which different physiological quantities were jointly encoded.
This validates the potential of self-supervision to capture important physiological information even in the absence of labeled data. 
Future work will have to demonstrate whether SSL can also be used successfully with other kinds of EEG recordings and tasks, such as regression.
Ultimately, developing a better understanding of how a pretext task can be designed to target a specific kind of EEG structure will be critical in establishing self-supervision as a key component of any EEG analysis pipeline.

\section*{Acknowledgements}
This work was supported by Mitacs (project number IT14765) and InteraXon Inc. (graduate funding support) for HB, by the ANR BrAIN grant for AG and DAE, and by a Fellowship from CIFAR and from the DATAIA convergence institute as part of the ``Programme d'Investissement d'Avenir'' (ANR-17-CONV-0003) operated by Inria for AH.

\bibliographystyle{unsrt}
\bibliography{bibli}  

\begin{thebibliography}{10}

\bibitem{ghassemi2018you}
Mohammad~M Ghassemi, Benjamin~E Moody, Li-Wei~H Lehman, Christopher Song, Qiao
  Li, Haoqi Sun, Roger~G Mark, M~Brandon Westover, and Gari~D Clifford.
\newblock You snooze, you win: the physionet/computing in cardiology challenge
  2018.
\newblock In {\em 2018 Computing in Cardiology Conference (CinC)}, volume~45,
  pages 1--4. IEEE, 2018.

\bibitem{acharya2013automated}
U~Rajendra Acharya, S~Vinitha Sree, G~Swapna, Roshan~Joy Martis, and Jasjit~S
  Suri.
\newblock Automated {EEG} analysis of epilepsy: a review.
\newblock {\em Knowledge-Based Systems}, 45:147--165, 2013.

\bibitem{lotte2018review}
Fabien Lotte, Laurent Bougrain, Andrzej Cichocki, Maureen Clerc, Marco Congedo,
  Alain Rakotomamonjy, and Florian Yger.
\newblock A review of classification algorithms for {EEG}-based brain--computer
  interfaces: a 10 year update.
\newblock {\em Journal of Neural Engineering}, 15(3):031005, 2018.

\bibitem{mihajlovic2015wearable}
Vojkan {Mihajlović}, Bernard {Grundlehner}, Ruud {Vullers}, and Julien
  {Penders}.
\newblock Wearable, wireless {EEG} solutions in daily life applications: What
  are we missing?
\newblock {\em IEEE Journal of Biomedical and Health Informatics}, 19(1):6--21,
  2015.

\bibitem{casson2019wearable}
Alexander~J Casson.
\newblock Wearable {EEG} and beyond.
\newblock {\em Biomedical Engineering Letters}, 9(1):53--71, 2019.

\bibitem{malhotra2013sleep}
Raman~K Malhotra and Alon~Y Avidan.
\newblock Sleep stages and scoring technique.
\newblock {\em Atlas of Sleep Medicine}, pages 77--99, 2013.

\bibitem{younes2016staging}
Magdy Younes, Jill Raneri, and Patrick Hanly.
\newblock Staging sleep in polysomnograms: Analysis of inter-scorer
  variability.
\newblock {\em Journal of Clinical Sleep Medicine}, 12(06):885--894, 2016.

\bibitem{engemann018robust}
Denis~A Engemann, Federico Raimondo, Jean-Rémi King, Benjamin Rohaut, Gilles
  Louppe, Frédéric Faugeras, Jitka Annen, Helena Cassol, Olivia Gosseries,
  Diego Fernandez-Slezak, Steven Laureys, Lionel Naccache, Stanislas Dehaene,
  and Jacobo~D Sitt.
\newblock Robust {EEG}-based cross-site and cross-protocol classification of
  states of consciousness.
\newblock {\em Brain}, 141(11):awy251, 2018.

\bibitem{jing2019self}
Longlong Jing and Yingli Tian.
\newblock Self-supervised visual feature learning with deep neural networks: A
  survey.
\newblock {\em arXiv preprint arXiv:1902.06162}, 2019.

\bibitem{noroozi2016unsupervised}
Mehdi Noroozi and Paolo Favaro.
\newblock Unsupervised learning of visual representations by solving jigsaw
  puzzles.
\newblock In {\em ECCV}, pages 69--84. Springer, 2016.

\bibitem{oord2018representation}
Aaron van~den Oord, Yazhe Li, and Oriol Vinyals.
\newblock Representation learning with contrastive predictive coding.
\newblock {\em arXiv preprint arXiv:1807.03748}, 2018.

\bibitem{mikolov2013efficient}
Tomas Mikolov, Kai Chen, Greg Corrado, and Jeffrey Dean.
\newblock Efficient estimation of word representations in vector space.
\newblock {\em arXiv preprint arXiv:1301.3781}, 2013.

\bibitem{devlin2018bert}
Jacob Devlin, Ming-Wei Chang, Kenton Lee, and Kristina Toutanova.
\newblock {BERT}: Pre-training of deep bidirectional transformers for language
  understanding.
\newblock {\em arXiv preprint arXiv:1810.04805}, 2018.

\bibitem{he2019rethinking}
Kaiming He, Ross Girshick, and Piotr Doll{\'a}r.
\newblock Rethinking imagenet pre-training.
\newblock In {\em Proceedings of the IEEE International Conference on Computer
  Vision}, pages 4918--4927, 2019.

\bibitem{yuan2017wave}
Ye~{Yuan}, Guangxu {Xun}, Qiuling {Suo}, Kebin {Jia}, and Aidong {Zhang}.
\newblock {Wave2Vec}: Learning deep representations for biosignals.
\newblock In {\em 2017 IEEE International Conference on Data Mining (ICDM)},
  pages 1159--1164, Nov 2017.

\bibitem{sarkar2020self}
Pritam Sarkar and Ali Etemad.
\newblock Self-supervised {ECG} representation learning for emotion
  recognition.
\newblock {\em arXiv preprint arXiv:2002.03898}, 2020.

\bibitem{roy2019deep}
Yannick Roy, Hubert Banville, Isabela Albuquerque, Alexandre Gramfort, Tiago~H
  Falk, and Jocelyn Faubert.
\newblock Deep learning-based electroencephalography analysis: a systematic
  review.
\newblock {\em Journal of Neural Engineering}, 16(5):051001, 2019.

\bibitem{engemann2020combining}
Denis~A Engemann, Oleh Kozynets, David Sabbagh, Guillaume Lema{\^\i}tre, Gael
  Varoquaux, Franziskus Liem, and Alexandre Gramfort.
\newblock Combining magnetoencephalography with magnetic resonance imaging
  enhances learning of surrogate-biomarkers.
\newblock {\em eLife}, 9:e54055, 2020.

\bibitem{obeid2016temple}
Iyad Obeid and Joseph Picone.
\newblock The temple university hospital {EEG} data corpus.
\newblock {\em Frontiers in Neuroscience}, 10:196, 2016.

\bibitem{bycroft2018uk}
Clare Bycroft, Colin Freeman, Desislava Petkova, Gavin Band, Lloyd~T Elliott,
  Kevin Sharp, Allan Motyer, Damjan Vukcevic, Olivier Delaneau, Jared
  O’Connell, et~al.
\newblock The {UK} {Biobank} resource with deep phenotyping and genomic data.
\newblock {\em Nature}, 562(7726):203--209, 2018.

\bibitem{shafto2014cambridge}
Meredith~A Shafto, Lorraine~K Tyler, Marie Dixon, Jason~R Taylor, James~B Rowe,
  Rhodri Cusack, Andrew~J Calder, William~D Marslen-Wilson, John Duncan, Tim
  Dalgleish, et~al.
\newblock The {Cambridge Centre for Ageing and Neuroscience (Cam-CAN)} study
  protocol: a cross-sectional, lifespan, multidisciplinary examination of
  healthy cognitive ageing.
\newblock {\em BMC neurology}, 14(1):204, 2014.

\bibitem{doersch2015unsupervised}
Carl Doersch, Abhinav Gupta, and Alexei~A Efros.
\newblock Unsupervised visual representation learning by context prediction.
\newblock In {\em ICCV}, pages 1422--1430, 2015.

\bibitem{misra2016shuffle}
Ishan Misra, C~Lawrence Zitnick, and Martial Hebert.
\newblock Shuffle and learn: unsupervised learning using temporal order
  verification.
\newblock In {\em ECCV}, pages 527--544. Springer, 2016.

\bibitem{turian2010word}
Joseph Turian, Lev Ratinov, and Yoshua Bengio.
\newblock Word representations: a simple and general method for semi-supervised
  learning.
\newblock In {\em Association for Computational Linguistics Annual Meeting},
  pages 384--394. Association for Computational Linguistics, 2010.

\bibitem{nayak2016evaluating}
Neha Nayak, Gabor Angeli, and Christopher~D Manning.
\newblock Evaluating word embeddings using a representative suite of practical
  tasks.
\newblock In {\em Proceedings of the 1st Workshop on Evaluating Vector-Space
  Representations for NLP}, pages 19--23, 2016.

\bibitem{henaff2019data}
Olivier~J H{\'e}naff, Ali Razavi, Carl Doersch, SM~Eslami, and Aaron van~den
  Oord.
\newblock Data-efficient image recognition with contrastive predictive coding.
\newblock {\em arXiv preprint arXiv:1905.09272}, 2019.

\bibitem{he2019momentum}
Kaiming He, Haoqi Fan, Yuxin Wu, Saining Xie, and Ross Girshick.
\newblock Momentum contrast for unsupervised visual representation learning.
\newblock {\em arXiv preprint arXiv:1911.05722}, 2019.

\bibitem{chen2020improved}
Xinlei Chen, Haoqi Fan, Ross Girshick, and Kaiming He.
\newblock Improved baselines with momentum contrastive learning.
\newblock {\em arXiv preprint arXiv:2003.04297}, 2020.

\bibitem{chen2020simple}
Ting Chen, Simon Kornblith, Mohammad Norouzi, and Geoffrey Hinton.
\newblock A simple framework for contrastive learning of visual
  representations.
\newblock {\em arXiv preprint arXiv:2002.05709}, 2020.

\bibitem{hyva17aistats}
Aapo Hyv\"arinen and Hiroshi Morioka.
\newblock Nonlinear {ICA} of temporally dependent stationary sources.
\newblock In {\em AISTATS}, 2017.

\bibitem{hyvarinen2019nonlinear}
Aapo Hyv\"arinen, Hiroaki Sasaki, and Richard~E Turner.
\newblock Nonlinear {ICA} using auxiliary variables and generalized contrastive
  learning.
\newblock In {\em AISTATS}, 2019.

\bibitem{makeig1997blind}
Scott Makeig, Tzyy-Ping Jung, Anthony~J Bell, Dara Ghahremani, and Terrence~J
  Sejnowski.
\newblock Blind separation of auditory event-related brain responses into
  independent components.
\newblock {\em Proceedings of the National Academy of Sciences},
  94(20):10979--10984, 1997.

\bibitem{jung1998extended}
Tzyy-Ping Jung, Colin Humphries, Te-Won Lee, Scott Makeig, Martin~J McKeown,
  Vicente Iragui, and Terrence~J Sejnowski.
\newblock Extended {ICA} removes artifacts from electroencephalographic
  recordings.
\newblock In {\em Advances in neural information processing systems}, pages
  894--900, 1998.

\bibitem{parra2005recipes}
Lucas~C Parra, Clay~D Spence, Adam~D Gerson, and Paul Sajda.
\newblock Recipes for the linear analysis of {EEG}.
\newblock {\em Neuroimage}, 28(2):326--341, 2005.

\bibitem{ablin2018faster}
Pierre Ablin, Jean-Fran{\c{c}}ois Cardoso, and Alexandre Gramfort.
\newblock Faster independent component analysis by preconditioning with
  {Hessian} approximations.
\newblock {\em IEEE Transactions on Signal Processing}, 66(15):4040--4049,
  2018.

\bibitem{becker1993learning}
Suzanna Becker.
\newblock Learning to categorize objects using temporal coherence.
\newblock In {\em Advances in neural information processing systems}, pages
  361--368, 1993.

\bibitem{wiskott2002slow}
Laurenz Wiskott and Terrence~J Sejnowski.
\newblock Slow feature analysis: Unsupervised learning of invariances.
\newblock {\em Neural Computation}, 14(4):715--770, 2002.

\bibitem{altevogt2006sleep}
Bruce~M Altevogt and Harvey~R Colten.
\newblock {\em Sleep disorders and sleep deprivation: an unmet public health
  problem}.
\newblock National Academies Press, 2006.

\bibitem{goldberger2000physiobank}
Ary~L Goldberger, Luis~AN Amaral, Leon Glass, Jeffrey~M Hausdorff, Plamen~Ch
  Ivanov, Roger~G Mark, Joseph~E Mietus, George~B Moody, Chung-Kang Peng, and
  H~Eugene Stanley.
\newblock {PhysioBank, PhysioToolkit, and PhysioNet}: components of a new
  research resource for complex physiologic signals.
\newblock {\em Circulation}, 101(23):e215--e220, 2000.

\bibitem{lopez2017automated}
Silvia L{\'o}pez, I~Obeid, and J~Picone.
\newblock Automated interpretation of abnormal adult electroencephalograms.
\newblock {\em MS Thesis, Temple University}, 2017.

\bibitem{bathgate2019diagnostic}
Christina~Jayne Bathgate and Jack~D Edinger.
\newblock Diagnostic criteria and assessment of sleep disorders.
\newblock In {\em Handbook of Sleep Disorders in Medical Conditions}, pages
  3--25. Elsevier, 2019.

\bibitem{chambon2018deep}
Stanislas Chambon, Mathieu~N Galtier, Pierrick~J Arnal, Gilles Wainrib, and
  Alexandre Gramfort.
\newblock A deep learning architecture for temporal sleep stage classification
  using multivariate and multimodal time series.
\newblock {\em IEEE Transactions on Neural Systems and Rehabilitation
  Engineering}, 26(4):758--769, 2018.

\bibitem{motamedi2014signal}
Shayan Motamedi-Fakhr, Mohamed Moshrefi-Torbati, Martyn Hill, Catherine~M.
  Hill, and Paul~R. White.
\newblock Signal processing techniques applied to human sleep {EEG} signals—a
  review.
\newblock {\em Biomedical Signal Processing and Control}, 10:21 -- 33, 2014.

\bibitem{smith2005eeg}
SJM Smith.
\newblock {EEG} in the diagnosis, classification, and management of patients
  with epilepsy.
\newblock {\em Journal of Neurology, Neurosurgery \& Psychiatry}, 76(suppl
  2):ii2--ii7, 2005.

\bibitem{micanovic2014diagnostic}
Christina Micanovic and Suvankar Pal.
\newblock The diagnostic utility of {EEG} in early-onset dementia: a systematic
  review of the literature with narrative analysis.
\newblock {\em Journal of Neural Transmission}, 121(1):59--69, 2014.

\bibitem{lopez2015automated}
S~Lopez, G~Suarez, D~Jungreis, I~Obeid, and J~Picone.
\newblock Automated identification of abnormal adult {EEG}s.
\newblock In {\em 2015 IEEE Signal Processing in Medicine and Biology Symposium
  (SPMB)}, pages 1--5. IEEE, 2015.

\bibitem{tibor2017deep}
Robin Tibor~Schirrmeister, Lukas Gemein, Katharina Eggensperger, Frank Hutter,
  and Tonio Ball.
\newblock Deep learning with convolutional neural networks for decoding and
  visualization of {EEG} pathology.
\newblock {\em arXiv preprint arXiv:1708.08012}, 2017.

\bibitem{gemein2020machine}
Lukas~AW Gemein, Robin~T Schirrmeister, Patryk Chrab{\k{a}}szcz, Daniel Wilson,
  Joschka Boedecker, Andreas Schulze-Bonhage, Frank Hutter, and Tonio Ball.
\newblock Machine-learning-based diagnostics of {EEG} pathology.
\newblock {\em Neuroimage}, page 117021, 2020.

\bibitem{ba2016layer}
Jimmy~Lei Ba, Jamie~Ryan Kiros, and Geoffrey~E Hinton.
\newblock Layer normalization.
\newblock {\em arXiv preprint arXiv:1607.06450}, 2016.

\bibitem{kingma2014adam}
Diederik~P Kingma and Jimmy Ba.
\newblock Adam: A method for stochastic optimization.
\newblock {\em arXiv preprint arXiv:1412.6980}, 2014.

\bibitem{he2015delving}
Kaiming He, Xiangyu Zhang, Shaoqing Ren, and Jian Sun.
\newblock Delving deep into rectifiers: Surpassing human-level performance on
  imagenet classification.
\newblock In {\em ICCV}, pages 1026--1034, 2015.

\bibitem{kramer1991nonlinear}
Mark~A Kramer.
\newblock Nonlinear principal component analysis using autoassociative neural
  networks.
\newblock {\em AIChE journal}, 37(2):233--243, 1991.

\bibitem{schirrmeister2017deep}
Robin~Tibor Schirrmeister, Jost~Tobias Springenberg, Lukas Dominique~Josef
  Fiederer, Martin Glasstetter, Katharina Eggensperger, Michael Tangermann,
  Frank Hutter, Wolfram Burgard, and Tonio Ball.
\newblock Deep learning with convolutional neural networks for {EEG} decoding
  and visualization.
\newblock {\em Human Brain Mapping}, aug 2017.

\bibitem{gramfort2014mne}
Alexandre Gramfort, Martin Luessi, Eric Larson, Denis~A Engemann, Daniel
  Strohmeier, Christian Brodbeck, Lauri Parkkonen, and Matti~S
  H{\"a}m{\"a}l{\"a}inen.
\newblock {MNE software for processing {MEG} and {EEG} data}.
\newblock {\em Neuroimage}, 86:446--460, 2014.

\bibitem{paszke2017automatic}
Adam Paszke, Sam Gross, Soumith Chintala, Gregory Chanan, Edward Yang, Zachary
  DeVito, Zeming Lin, Alban Desmaison, Luca Antiga, and Adam Lerer.
\newblock Automatic differentiation in {PyTorch}.
\newblock In {\em NIPS}, 2017.

\bibitem{BARACHANT2013172}
Alexandre Barachant, Stéphane Bonnet, Marco Congedo, and Christian Jutten.
\newblock Classification of covariance matrices using a {Riemannian}-based
  kernel for {BCI} applications.
\newblock {\em Neurocomputing}, 112:172 -- 178, 2013.
\newblock Advances in artificial neural networks, machine learning, and
  computational intelligence.

\bibitem{pedregosa2011scikit}
Fabian Pedregosa, Ga{\"e}l Varoquaux, Alexandre Gramfort, Vincent Michel,
  Bertrand Thirion, Olivier Grisel, Mathieu Blondel, Peter Prettenhofer, Ron
  Weiss, Vincent Dubourg, et~al.
\newblock {Scikit-learn: Machine learning in Python}.
\newblock {\em Journal of Machine Learning Research}, 12(Oct):2825--2830, 2011.

\bibitem{berry2012aasm}
Richard~B Berry, Rita Brooks, Charlene~E Gamaldo, Susan~M Harding, Carole~L
  Marcus, Bradley~V Vaughn, et~al.
\newblock The {AASM} manual for the scoring of sleep and associated events.
\newblock {\em Rules, Terminology and Technical Specifications, American
  Academy of Sleep Medicine}, 176, 2012.

\bibitem{aboalayon2016sleep}
Khald Aboalayon, Miad Faezipour, Wafaa Almuhammadi, and Saeid Moslehpour.
\newblock Sleep stage classification using {EEG} signal analysis: a
  comprehensive survey and new investigation.
\newblock {\em Entropy}, 18(9):272, 2016.

\bibitem{mcinnes2018umap}
Leland McInnes, John Healy, and James Melville.
\newblock Umap: Uniform manifold approximation and projection for dimension
  reduction.
\newblock {\em arXiv preprint arXiv:1802.03426}, 2018.

\bibitem{pardey1996new}
James Pardey, Stephen Roberts, Lionel Tarassenko, and John Stradling.
\newblock A new approach to the analysis of the human sleep/wakefulness
  continuum.
\newblock {\em Journal of sleep research}, 5(4):201--210, 1996.

\bibitem{lopour2011continuous}
Beth~A Lopour, Savas Tasoglu, Heidi~E Kirsch, James~W Sleigh, and Andrew~J
  Szeri.
\newblock A continuous mapping of sleep states through association of {EEG}
  with a mesoscale cortical model.
\newblock {\em Journal of computational neuroscience}, 30(2):471--487, 2011.

\bibitem{mander2017sleep}
Bryce~A Mander, Joseph~R Winer, and Matthew~P Walker.
\newblock Sleep and human aging.
\newblock {\em Neuron}, 94(1):19--36, 2017.

\bibitem{purcell2017characterizing}
SM~Purcell, DS~Manoach, C~Demanuele, BE~Cade, S~Mariani, R~Cox,
  G~Panagiotaropoulou, R~Saxena, JQ~Pan, JW~Smoller, et~al.
\newblock Characterizing sleep spindles in 11,630 individuals from the national
  sleep research resource.
\newblock {\em Nature Communications}, 8:15930, 2017.

\bibitem{woodman2010brief}
Geoffrey~F Woodman.
\newblock A brief introduction to the use of event-related potentials in
  studies of perception and attention.
\newblock {\em Attention, Perception, \& Psychophysics}, 72(8):2031--2046,
  2010.

\bibitem{abiri2019comprehensive}
Reza Abiri, Soheil Borhani, Eric~W Sellers, Yang Jiang, and Xiaopeng Zhao.
\newblock A comprehensive review of {EEG}-based brain--computer interface
  paradigms.
\newblock {\em Journal of Neural Engineering}, 16(1):011001, 2019.

\bibitem{banville2019self}
Hubert Banville, Graeme Moffat, Isabela Albuquerque, Denis-Alexander Engemann,
  Aapo Hyv{\"a}rinen, and Alexandre Gramfort.
\newblock Self-supervised representation learning from electroencephalography
  signals.
\newblock In {\em 2019 IEEE 29th International Workshop on Machine Learning for
  Signal Processing (MLSP)}, pages 1--6. IEEE, 2019.

\bibitem{tian2020makes}
Yonglong Tian, Chen Sun, Ben Poole, Dilip Krishnan, Cordelia Schmid, and
  Phillip Isola.
\newblock What makes for good views for contrastive learning, 2020.

\bibitem{loomis1937cerebral}
Alfred~L Loomis, E~Newton Harvey, and Garret~A Hobart.
\newblock Cerebral states during sleep, as studied by human brain potentials.
\newblock {\em Journal of experimental psychology}, 21(2):127, 1937.

\bibitem{kales1968manual}
Anthony Kales, Allan Rechtschaffen, Los Angeles. Brain Information~Service
  University~of California, and NINDB Neurological Information~Network (U.S.).
\newblock {\em A Manual of Standardized Terminology, Techniques and Scoring
  System for Sleep Stages of Human Subjects: Allan Rechtschaffen and Anthony
  Kales, Editors}.
\newblock NIH publication. U. S. National Institute of Neurological Diseases
  and Blindness, Neurological Information Network, 1968.

\bibitem{moser2009sleep}
Doris Moser, Peter Anderer, Georg Gruber, Silvia Parapatics, Erna Loretz,
  Marion Boeck, Gerhard Kloesch, Esther Heller, Andrea Schmidt, Heidi
  Danker-Hopfe, et~al.
\newblock Sleep classification according to {AASM} and {Rechtschaffen} \&
  {Kales}: effects on sleep scoring parameters.
\newblock {\em Sleep}, 32(2):139--149, 2009.

\bibitem{schulz2008rethinking}
Hartmut Schulz.
\newblock Rethinking sleep analysis comment on the aasm manual for the scoring
  of sleep and associated events.
\newblock {\em Journal of Clinical Sleep Medicine}, 4(02):99--103, 2008.

\bibitem{pack2019epilepsy}
Alison~M Pack.
\newblock Epilepsy overview and revised classification of seizures and
  epilepsies.
\newblock {\em CONTINUUM: Lifelong Learning in Neurology}, 25(2):306--321,
  2019.

\bibitem{england2012epilepsy}
Mary~Jane England, Catharyn~T Liverman, Andrea~M Schultz, and Larisa~M
  Strawbridge.
\newblock Epilepsy across the spectrum: Promoting health and understanding.: A
  summary of the institute of medicine report.
\newblock {\em Epilepsy \& Behavior}, 25(2):266--276, 2012.

\bibitem{walters2010dementia}
Glenn~D Walters.
\newblock Dementia: Continuum or distinct entity?
\newblock {\em Psychology and aging}, 25(3):534, 2010.

\bibitem{supratak2017deepsleepnet}
Akara Supratak, Hao Dong, Chao Wu, and Yike Guo.
\newblock {DeepSleepNet}: a model for automatic sleep stage scoring based on
  raw single-channel {EEG}.
\newblock {\em IEEE Transactions on Neural Systems and Rehabilitation
  Engineering}, 25(11):1998--2008, 2017.

\bibitem{hyvarinen2016unsupervised}
Aapo Hyv{\"a}rinen and Hiroshi Morioka.
\newblock Unsupervised feature extraction by time-contrastive learning and
  nonlinear {ICA}.
\newblock In {\em Advances in Neural Information Processing Systems}, pages
  3765--3773, 2016.

\bibitem{kolesnikov2019revisiting}
Alexander Kolesnikov, Xiaohua Zhai, and Lucas Beyer.
\newblock Revisiting self-supervised visual representation learning.
\newblock In {\em Proceedings of the IEEE Conference on Computer Vision and
  Pattern Recognition}, pages 1920--1929, 2019.

\end{thebibliography}

\appendix

\section{UMAP on RP-learned features}

\begin{figure}[h]
    \centering
    \includegraphics[width=\textwidth]{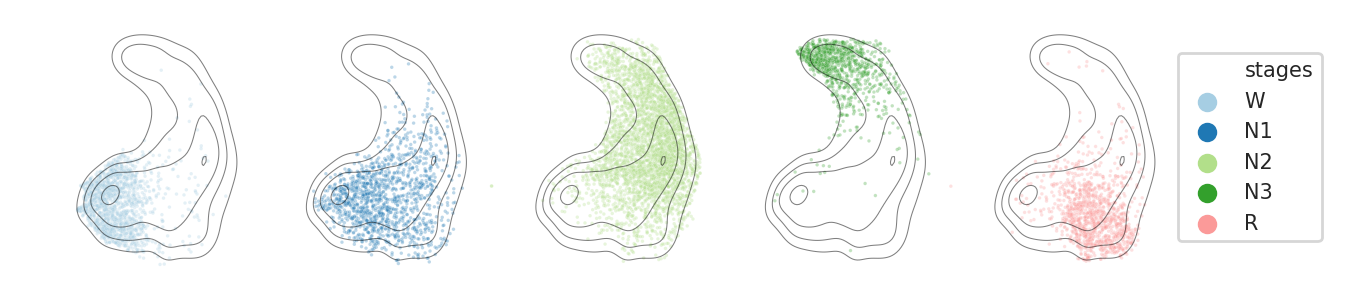}
    \caption{UMAP visualization of RP-learned features on the entire PC18 dataset. See Fig.~\ref{fig:umaps_stages} for a complete description.}
    \label{fig:umaps_stages_rp}
\end{figure}

\begin{figure}[h]
    \centering
    \includegraphics[width=\textwidth]{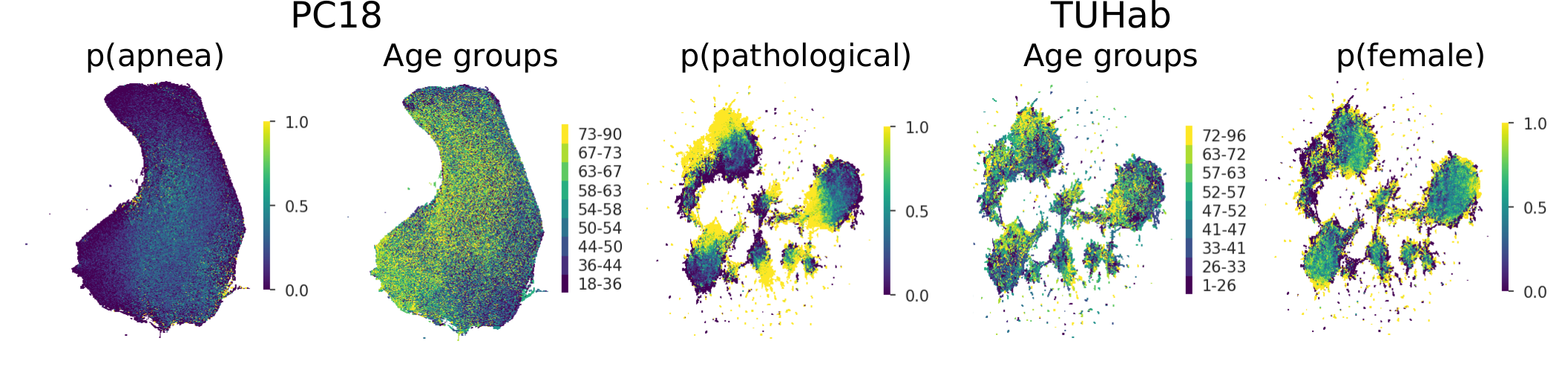}
    \caption{Structure learned by the embedders trained on the RP task. See Fig.~\ref{fig:umaps_others} for a complete description.}
    \label{fig:umaps_others_rp}
\end{figure}

We show the results of the experiment of Section~\ref{subsec:umap_exp} for features learned with the RP pretext task in Figures~\ref{fig:umaps_stages_rp} and \ref{fig:umaps_others_rp}. 

\section{Hyperparameter search procedure}
\label{subsec:hp_search_procedure}

\begin{table}[h]
\begin{tabular}{lp{3.5cm}p{3cm}|p{3.5cm}p{3cm}}
    \toprule 
   & \multicolumn{2}{c|}{\textbf{PC18}}                                          & \multicolumn{2}{c}{\textbf{TUHab}}                                          \\
   \midrule
   & $\tau_{pos}$ and $\tau_{neg}$ (min) & $\tau_{neg}$ (min)          & $\tau_{pos}$ and $\tau_{neg}$ (min) & $\tau_{neg}$ (min)           \\
   \midrule
RP  & 0.5, 1, 2, 5, 15, 30, 60, 120 & 0, 0.5, 1, 2, 5, 15, 30, 60, 120  & 6 s, 12 s, 0.5, 1, 2, 5, 10 & 0 s, 6 s, 12 s, 0.5, 1, 2, 5, 10  \\
TS & 1, 2, 5, 15, 30, 60, 120      & 2, 5, 15, 30, 60, 120 & 12 s, 0.5, 1, 2, 5, 10        & 12 s, 0.5, 1, 2, 5, 10 \\
\\
\midrule
   & \# predicted windows         & Negative sampling           & \# predicted windows         & Negative sampling            \\
   \midrule
CPC & 2, 4, 8, 12, 16               & same recording, across recordings & 2, 4, 8, 12, 16             & same recording, across recordings \\
\bottomrule
\end{tabular}
\caption{SSL pretext task hyperparameter values considered in Experiment 2.}
\label{tab:hyperparams}
\end{table}

The hyperparameter search in Section~\ref{subsec:exp_hyperparameters} was carried out using the following steps.
First, embedders $h_\Theta$ were independently trained on the RP, TS and CPC tasks.
The parameters of $h_\Theta$ were then frozen, and the different $h_\Theta$ were used as feature extractors to obtain sets of 100-dimensional feature vectors from the original input data.
Finally, we trained linear logistic regression classifiers to perform the downstream tasks given the extracted features.
We further varied the principal pretext task hyperparameters to understand their impact on both pretext and downstream task performance (see Table~\ref{tab:hyperparams}).
In both cases, we compared the balanced accuracy on the validation set.
For RP and TS, we focused our attention on $\tau_{pos}$ and $\tau_{neg}$, which are used to control the size of the positive and negative contexts when sampling pairs or triplets of windows.
As a first step, the values of $\tau_{pos}$ and $\tau_{neg}$ were varied jointly, i.e., $\tau_{pos} = \tau_{neg}$, to avoid sampling ``confusing'' pairs or triplets of windows which could come from either the positive or negative classes.
The best value was then used to set $\tau_{pos}$, and a sweep over different $\tau_{neg}$ values was carried out.
In a second step, we fixed $\tau_{neg}$ such that it encompassed all recordings, i.e., negative windows were uniformly sampled from any recording in the dataset instead of being limited to the recording which contains the anchor window.
We then again varied $\tau_{pos}$ with this second negative sampling strategy.
For CPC, we studied the impact of the number of predicted windows (``\# steps'') \cite{oord2018representation} and, as for RP and TS, the type of negative sampling (``same-recording'' vs. ``across-recordings'').
Again, we first varied the number of predicted windows and reused the best value to compare negative sampling strategies.

\end{document}